\documentclass[journal]{IEEEtran}
\usepackage{amsmath, amssymb}

\usepackage{graphicx}
\usepackage{caption}
\usepackage{subcaption}
\usepackage{color}

\usepackage{hyperref}

\usepackage{csquotes}

\usepackage{url}

\usepackage{algorithm}

\usepackage{multirow}

\usepackage{bm}

\begin{document}

\title{Context-adaptive neural network based prediction for image compression}

\author{Thierry Dumas, Aline Roumy, Christine Guillemot
\thanks{Thierry Dumas is currently with Interdigital, 35576 Cesson-S{\'e}vign{\'e}, France (e-mail: thierry.dumas@interdigital.com). Aline Roumy and Christine Guillemot are with INRIA Rennes, 35042 Rennes, France (e-mail: aline.roumy@inria.fr, christine.guillemot@inria.fr).}
\thanks{This work has been supported by the French Defense Procurement Agency (DGA).}}

\maketitle


\begin{abstract} \label{sec:0}
This paper describes a set of neural network architectures, called Prediction Neural Networks Set (PNNS), based on both fully-connected and convolutional neural networks, for intra image prediction. The choice of neural network for predicting a given image block depends on the block size, hence does not need to be signalled to the decoder. It is shown that, while fully-connected neural networks give good performance for small block sizes, convolutional neural networks provide better predictions in large blocks with complex textures. Thanks to the use of masks of random sizes during training, the neural networks of PNNS well adapt to the available context that may vary, depending on the position of the image block to be predicted. When integrating PNNS into a H.265 codec, PSNR-rate performance gains going from $1.46\%$ to $5.20\%$ are obtained. These gains are on average $0.99\%$ larger than those of prior neural network based methods. Unlike the H.265 intra prediction modes, which are each specialized in predicting a specific texture, the proposed PNNS can model a large set of complex textures.
\end{abstract}

\begin{IEEEkeywords}
Image compression, intra prediction, neural networks.
\end{IEEEkeywords}


\section{Introduction} \label{sec:1}
\IEEEPARstart{I}{ntra} prediction is a key component of image and video compression algorithms and in particular of recent coding standards such as H.265 \cite{high_efficiency_video}. The goal of intra prediction is to infer a block of pixels from the previously encoded and decoded neighborhood. The predicted block is subtracted from the original block to yield a residue which is then encoded. Intra prediction modes used in practice rely on very simple models of dependencies between the block to be predicted and its neighborhood. This is the case of the H.265 standard which selects according to a rate-distortion criterion one mode among 35 fixed and simple prediction functions. The H.265 prediction functions consist in simply propagating the pixel values along specified directions \cite{intra_coding_of}. This approach is suitable in the presence of contours, hence in small regions containing oriented edges \cite{edges_are_the, independent_component_filters, sparse_deep_belief}. However, it fails in large areas usually containing more complex textures \cite{a_hierarchical_statistical, convolutional_deep_belief, modeling_natural_images}. Instead of simply propagating pixels in the causal neighborhood, the authors in \cite{intra_prediction_by} look for the best predictor within the image 
by searching for the best match with the so-called template of the block to be predicted. The authors in \cite{image_predition_based} further exploit self-similarities within the image with more complex models defined as linear combinations of k-nearest patches in the neighborhood.

In this paper, we consider the problem of designing an intra prediction function that can predict both simple textures in small image blocks, as well as complex textures in larger ones. To create an optimal intra prediction function, the probabilistic model of natural images is needed. Let us consider a pixel, denoted by the random variable $X$, to be predicted from its neighboring decoded pixels. These neighboring decoded pixels are represented as a set $\mathcal{B}$ of observed random variables. The authors in \cite{source_coding_part_I} demonstrate that the optimal prediction $\hat{X}^{*}$ of $X$, i.e. the prediction that minimizes the mean squared prediction error, is the conditional expectation $\mathbb{E} \left[ X \vert \mathcal{B} \right]$. Yet, no existing model of natural images gives a reliable $\mathbb{E} \left[ X \vert \mathcal{B} \right]$.

However, neural networks have proved capable of learning a reliable model of the probability of image pixels for prediction. For example, in \cite{generative_image_modeling, pixel_recurrent_neural}, recurrent neural networks sequentially update their internal representation of the dependencies between the pixels in the known region of an image and then generate the next pixel in the unknown region of the image.

In this paper, we consider the problem of learning, with the help of neural networks, a reliable model of dependencies between a block, possibly containing a complex texture, and its neighborhood that we refer to as its context. Note that neural networks have already been considered in \cite{fully_connected_network} for intra block prediction. However, the authors in \cite{fully_connected_network} only take into consideration blocks of sizes $4 \times 4$, $8 \times 8$, $16 \times 16$, and $32 \times 32$ pixels and use fully-connected neural networks. Here, we consider both fully-connected and convolutional neural networks. We show that, while fully-connected neural networks give good performance for small block sizes, convolutional neural networks are more appropriate, both in terms of prediction PSNR and PSNR-rate performance gains, for large block sizes. The choice of neural network is block size dependent, hence does not need to be signalled to the decoder. This set of neural networks, called Prediction Neural Networks Set (PNNS), has been integrated into a H.265 codec, showing PSNR-rate performance gains from $1.46\%$ to $5.20\%$.

In summary, the contributions of this paper are as follows:
\begin{itemize}
	\item We propose a set of neural network architectures, including both fully-connected and convolutional neural networks, for intra image prediction. 
	\item We show that, in the case of large block sizes, convolutional neural networks yield more accurate predictions compared with fully-connected ones.
	\item Thanks to the use of masks of random sizes during training, the neural networks of PNNS well adapt to the available context that may vary. E.g. in H.265, the available context, hence the number of known pixels in the neighborhood, depends on the position of the considered prediction unit within the coding unit and within the coding tree unit.
	\item Unlike the H.265 intra prediction modes, which are each specialized in predicting a specific texture, the proposed PNNS, trained on a large unconstrained set of images, is able to model a large set of complex textures.
	\item We prove experimentally a surprising property of the neural networks for intra prediction: they do not need to be trained on distorted contexts, meaning that the neural networks trained on undistorted contexts generalize well on distorted contexts, even for severe distortions.
\end{itemize}
The code to reproduce our numerical results and train the neural networks is available online \footnote{\url{https://github.com/thierrydumas/context_adaptive_neural_network_based_prediction} \label{footnote:website_code}}.


\section{Conditions for efficient neural network based intra prediction} \label{sec:2}
Prediction is a key method in rate distortion theory, when complexity is an issue. Indeed, the complexity of vector quantization is prohibitive, and scalar quantization is rather used. But, scalar quantization cannot exploit the statistical correlations between data samples. This task can be done via prediction \cite{quantization}. Prediction can only be made from data samples available at the decoder, i.e. causal and distorted data samples. By distorted causal data samples we mean previously encoded and decoded pixels above and on the left side of the image block to be predicted. This set of pixels is often referred to as the context of the block to be predicted.

Optimal prediction, i.e. conditional expectation \cite{source_coding_part_I}, requires knowing the conditional distribution of the image block to be predicted given causal and distorted data samples. Estimating such a conditional distribution is difficult. The use of the predictor by the decoder would in addition require sending the distribution parameters. Classical approaches in predictive coding consist in proposing a set of predefined functions and choosing the best of them in a rate-distortion sense. Thus, the number of possible functions is limited. On the other hand, neural networks can approximate many functions, in particular complex predictive functions such as the generation of future video frames given an input sequence of frames \cite{video_language_modeling, deep_multi_scale}.

But, the use of neural networks for intra prediction within an image coding scheme raises several questions that we address in this paper. What neural network architecture provides enough power of representation to map causal and distorted data samples to an accurate prediction of a given image block? What context size should be used? Section \ref{sec:3} looks for a neural network architecture and the optimal number of causal and distorted data samples for predicting a given image block. Moreover, the amount of causal and distorted data samples available at the decoder varies. It depends on the partitioning of the image and the position of the block to be predicted within the image. Section \ref{sec:4} trains the neural networks so that they adapt to the variable context size. Finally, can neural networks compensate for the quantization noise in its input and be efficient in a rate-distortion sense? Sections \ref{sec:5} and \ref{sec:6} answer these two questions with experimental evidence.


\section{Proposed neural network based intra prediction} \label{sec:3}
Unlike standard intra prediction in which the encoder chooses the best mode in a rate-distortion sense among several pre-defined modes, only one neural network among a set of neural networks does the prediction here. Unlike \cite{fully_connected_network}, our set contains both fully-connected and convolutional neural networks. This section first presents our set of neural networks. Then, it explains how one neural network is selected for predicting a given image block and the context is defined according to the block size. Finally, the specificities of the integration of our neural networks into H.265 are detailed.

\subsection{Fully-connected and convolutional neural networks} \label{subsec:3.1}
\begin{figure}
	\centering
	\includegraphics[width=0.22\textwidth]{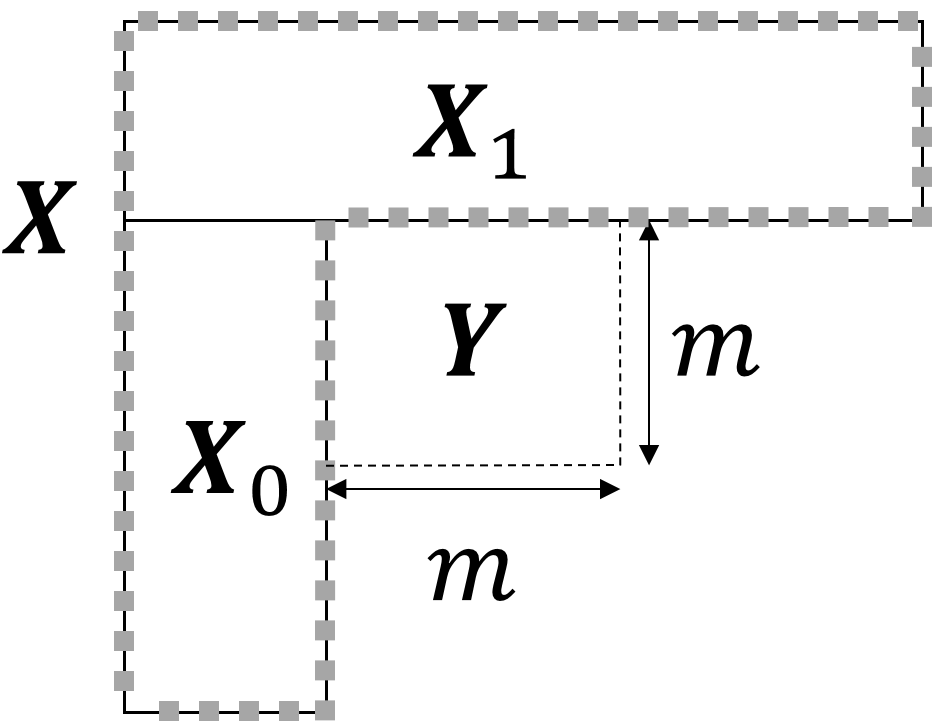}
	\caption{Illustration of the relative positions of $\mathbf{X}$, $\mathbf{X}_{0}$, $\mathbf{X}_{1}$, and $\mathbf{Y}$.}
	\label{fig:section3.1}
\end{figure}

\begin{figure}
	\centering
	\includegraphics[width=0.40\textwidth]{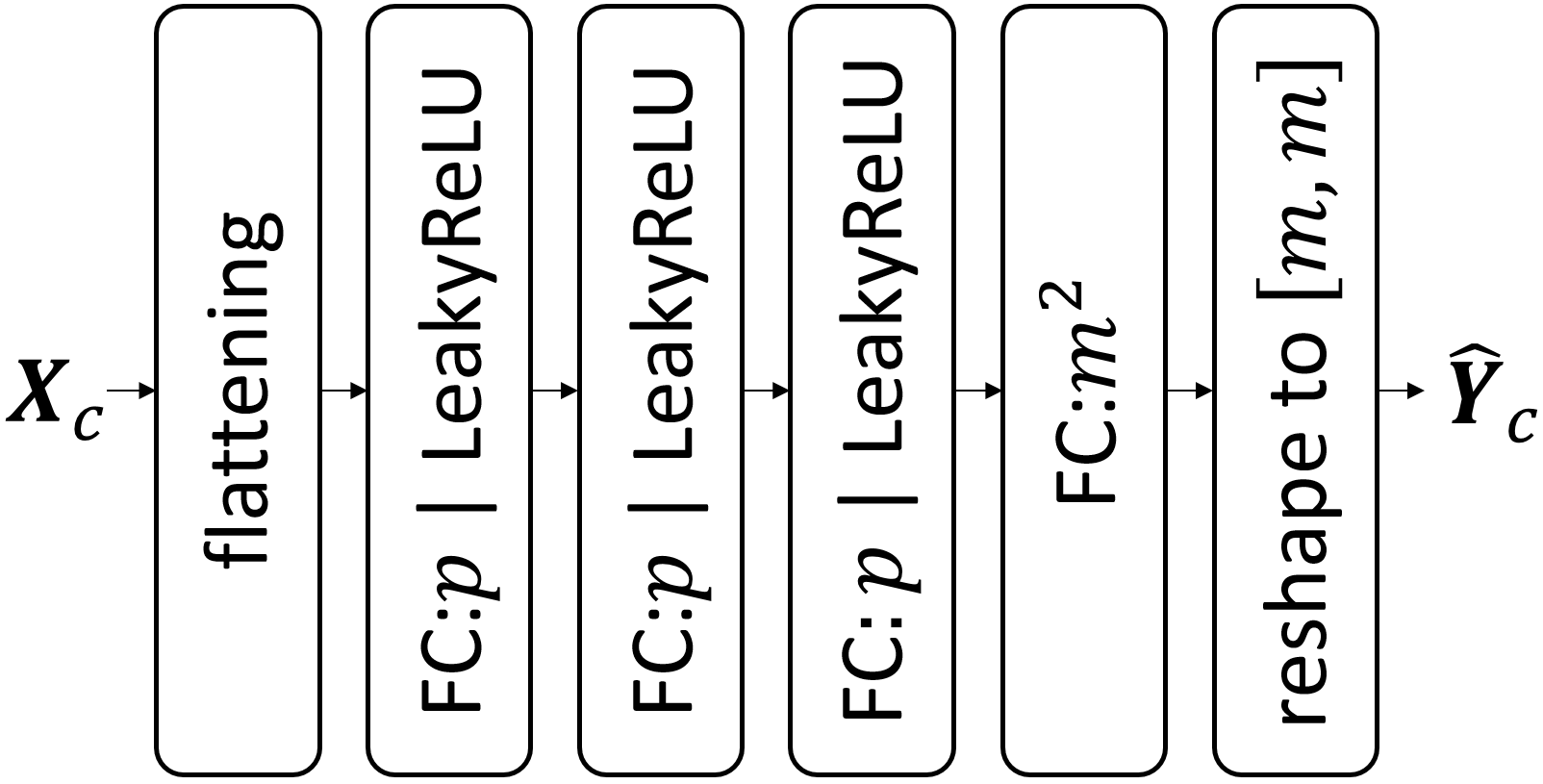}
	\caption{Illustration of the fully-connected architecture $f_{m}$. FC:$p \vert \text{LeakyReLU}$ denotes the fully-connected layer with $p$ output neurons and LeakyReLU with slope $0.1$ as non-linear activation. Unlike FC:$p \vert \text{LeakyReLU}$, FC:$p$ has no non-linear activation. $\boldsymbol{\theta}_{m}$ gathers the weights and biases of the $4$ fully-connected layers of $f_{m}$.}
	\label{fig:section3.2}
\end{figure}

\begin{figure*}
	\centering
	\includegraphics[width=0.80\textwidth]{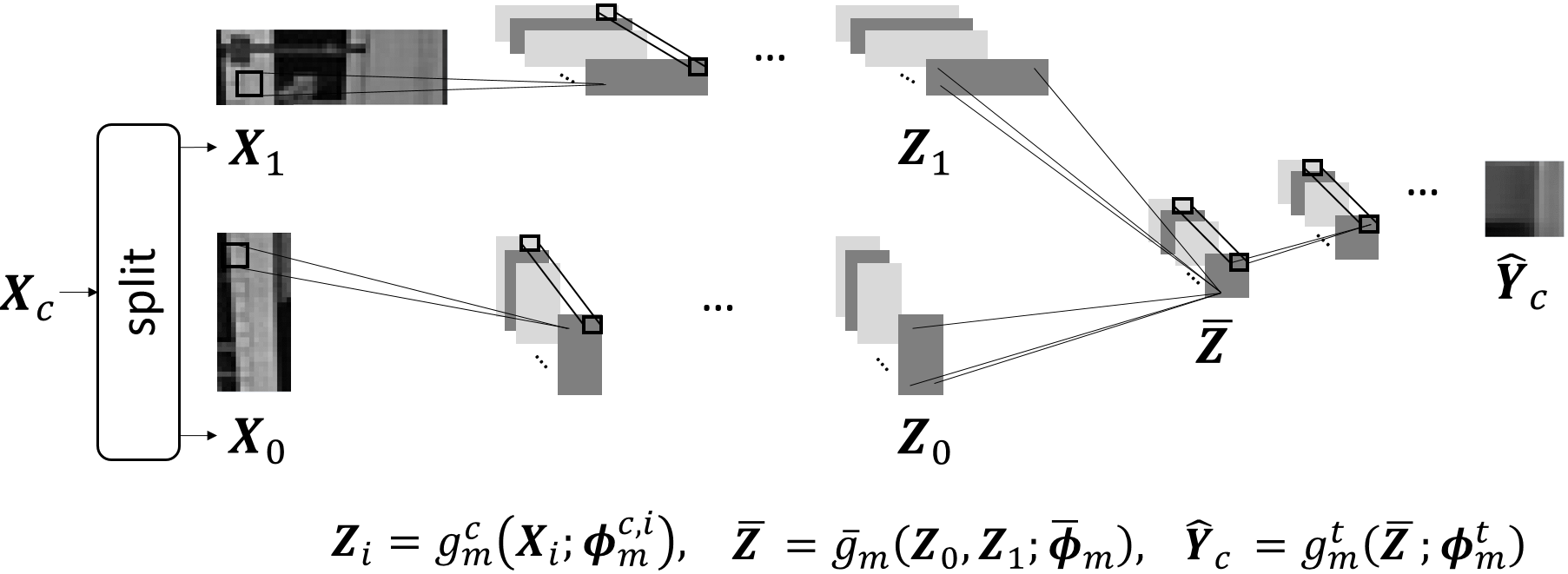}
	\caption{Illustration of the convolutional architecture $g_{m}$ in which $g_{m}^{c}$, $\overline{g}_{m}$, and $g_{m}^{t}$ denote respectively a stack of convolutional layers, the merger, and the stack of transposed convolutional layers. $\boldsymbol{\phi}_{m}^{c}$, $\overline{\boldsymbol{\phi}}_{m}$, and $\boldsymbol{\phi}_{m}^{t}$ gather the weights and biases of respectively $g_{m}^{c}$, $\overline{g}_{m}$, and $g_{m}^{t}$. $\boldsymbol{\phi}_{m} = \left\{ \boldsymbol{\phi}_{m}^{c, 0}, \boldsymbol{\phi}_{m}^{c, 1}, \overline{\boldsymbol{\phi}}_{m}, \boldsymbol{\phi}_{m}^{t} \right\}$.}
	\label{fig:section3.3}
\end{figure*}

Let $\mathbf{X}$ be a context containing decoded pixels above and on the left side of a square image block $\mathbf{Y}$ of width $m \in \mathbb{N}^{*}$ to be predicted (see Figure \ref{fig:section3.1}). The transformation of $\mathbf{X}$ into a prediction $\hat{\mathbf{Y}}$ of $\mathbf{Y}$ via either a fully-connected neural network $f_{m}$, parametrized by $\boldsymbol{\theta}_{m}$, or a convolutional neural network $g_{m}$, parametrized by $\boldsymbol{\phi}_{m}$, is described in \eqref{eq:1}. The corresponding architectures are depicted in Figures \ref{fig:section3.2} and \ref{fig:section3.3}.
\begin{align} \label{eq:1}
	&\mathbf{X}_{c} = \mathbf{X} - \alpha \nonumber\\
	&\hat{\mathbf{Y}}_{c} = \begin{cases} f_{m} \left( \mathbf{X}_{c}; \boldsymbol{\theta}_{m} \right) \\ g_{m} \left( \mathbf{X}_{c}; \boldsymbol{\phi}_{m} \right) \end{cases}\\
	&\hat{\mathbf{Y}} = \max \left( \min \left( \hat{\mathbf{Y}}_{c} + \alpha, 255 \right), 0 \right) \nonumber
\end{align}

During optimization, each input variable to a neural network must be approximatively zero-centered over the training set to accelerate convergence \cite{efficient_backprop}. Besides, since the pixel space corresponds to both the input and output spaces in intra prediction, it makes sense to normalize the input and output similarly. One could subtract from each image block to be predicted and its context their respective mean during training. But, this entails sending the mean of a block to be predicted to the decoder during the test phase. Instead, the mean pixel intensity $\alpha$ is first computed over all the training images. Then, $\alpha$ is subtracted from each image block to be predicted and its context during training. During the test phase, $\alpha$ is subtracted from the context (see \eqref{eq:1} where the subscript $c$ stands for centered).

This preprocessing implies a postprocessing of the neural network output. More precisely, the learned mean pixels intensity is added to the output and the result is clipped to $\left[ 0, 255 \right]$ (see \eqref{eq:1}).

The first operation for both architectures consists in formatting the context to ease the computations in the neural network. In the case of a fully-connected neural network, all elements in the context are connected such that there is no need to keep the 2D structure of the context \cite{gradient_based_learning}. Therefore, the context is first vectorized, and fast vector-matrix arithmetics can be used (see Figure \ref{fig:section3.2}). However, in the case of a convolutional neural network, fast computation of 2D filtering requires to keep the 2D structure of the context. Moreover, again for fast computation, the shape of the input to the convolution has to be rectangular. That is why the context is split into two rectangles $\mathbf{X}_{0}$ and $\mathbf{X}_{1}$ (see Figures \ref{fig:section3.1} and \ref{fig:section3.3}). $\mathbf{X}_{0}$ and $\mathbf{X}_{1}$ are then processed by distinct convolutions.

The proposed fully-connected architecture is composed of $4$ fully-connected layers. The first layer computes an overcomplete representation of the context to reach $p \in \mathbb{N}^{*}$ output coefficients. Overcompleteness is chosen as it is observed empirically that overcomplete representations in early layers boost the performance of neural networks \cite{reducing_the_dimensionality, using_very_deep, very_deep_convolutional}. The next two layers keep the number of coefficients unchanged, while the last layer reduces it to provide the predicted image block. The first three layers have LeakyReLU \cite{rectifier_nonlinearities_improve} with slope $0.1$ as non-linear activation. The last layer has no non-linear activation. This is because the postprocessing discussed earlier contains already a non-linearity, which consists in first adding the learned mean pixels intensity to the output and clipping the result to $\left[ 0, 255 \right]$.

The first task of the convolutional architecture is the computation of features characterizing the dependencies between the elements in $\mathbf{X}_{0}$. $\mathbf{X}_{0}$ is thus fed into a stack of convolutional layers. This yields a stack $\mathbf{Z}_{0}$ of $l \in \mathbb{N}^{*}$ feature maps (see Figure \ref{fig:section3.3}). Similarly, $\mathbf{X}_{1}$ is fed into another stack of convolutional layers. This yields a stack $\mathbf{Z}_{1}$ of $l$ feature maps.

All the elements in the context can be relevant for predicting any image block pixel. This implies that the information associated to all spatial locations in the context has to be merged. That is why the next layer in the convolutional architecture merges spatially $\mathbf{Z}_{0}$ and $\mathbf{Z}_{1}$ (see Figure \ref{fig:section3.3}). More precisely, for $i \in \left[ \left\vert 1, l \right\vert \right]$, all the coefficients of the $i^{\text{th}}$ feature map of $\mathbf{Z}_{0}$ and of the $i^{\text{th}}$ feature map of $\mathbf{Z}_{1}$ are merged through affine combinations. Then, LeakyReLU with slope $0.1$ is applied, yielding the merged stack $\overline{\mathbf{Z}}$ of feature maps. Note that this layer bears similarities with the \enquote{channelwise fully-connected layer} \cite{context_encoders_feature}. But, unlike the \enquote{channelwise fully-connected layer}, it merges two stacks of feature maps of different height and width. Its advantage over a fully-connected layer is that it contains $l$ times less weights.

The last task of the convolutional architecture is to merge the information of the different feature maps of $\overline{\mathbf{Z}}$. $\overline{\mathbf{Z}}$ is thus fed into a stack of transposed convolutional layers \cite{learning_deconvolution_network, a_guide_to}. This yields the predicted image block (see Figure \ref{fig:section3.3}). Note that all convolutional layers and transposed convolutional layers, apart from the last transposed convolutional layer, have LeakyReLU with slope $0.1$ as non-linear activation. The last transposed convolutional layer has no non-linear activation due to the postprocessing discussed earlier.

\subsection{Growth rate of the context size with the block size} \label{subsec:3.2}
Now that the architectures and the shape of the context are defined, the size of the context remains to be optimized. The causal neighborhood of the image block to be predicted used by the H.265 intra prediction modes is limited to one row of $2m + 1$ decoded pixels above the block and one column of $2m$ decoded pixels on the left side of the block. However, a context of such a small size is not sufficient for neural networks as a neural network relies on the spatial distribution of the decoded pixels intensity in the context to predict complex textures. Therefore, the context size has to be larger than $4m + 1$.

But, an issue arises when the size of the context grows too much. Indeed, if the image block to be predicted is close to either the top edge of the decoded image $\mathbf{D}$ or its left edge, a large context goes out of the bounds of the decoded image. The neural network prediction is impossible. There is thus a tradeoff to find a suitable size for the context.

\begin{figure}
	\centering
	\includegraphics[width=0.48\textwidth]{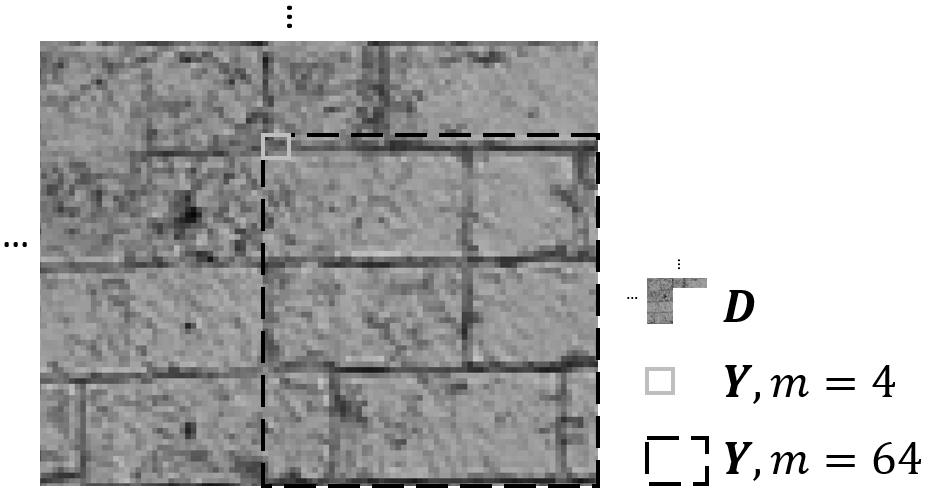}
	\caption[caption]{Dependencies between $\mathbf{D}$ and $\mathbf{Y}$. The luminance channel of the first image in the Kodak suite \cite{website_kodak} is being encoded via H.265 with Quantization Parameter $\text{QP} = 17$.}
	\label{fig:section3.4}
\end{figure}

Let us look at decoded pixels above and on the left side of $\mathbf{Y}$ to develop intuitions regarding this tradeoff. When $m$ is small, the long range spatial dependencies between these decoded pixels are not relevant for predicting $\mathbf{Y}$ (see Figure \ref{fig:section3.4}). In this case, the size of the context should be small so that the above-mentioned issue is limited. However, when $m$ is large, such long range spatial dependencies are informative for predicting $\mathbf{Y}$. The size of the context should now be large, despite the issue. Therefore, the context size should be a function of $m^{q}, q \geq 1$.

From there, we ran several preliminary experiments in which $q \in \left\{ 1, 2, 3 \right\}$ and the PSNR between $\hat{\mathbf{Y}}$ and $\mathbf{Y}$ was measured. The conclusion is that a neural network yields the best PSNRs when the size of the context grows with $m^{2}$, i.e. the ratio between the size of the context and the size of the image block to be predicted is constant. This makes sense as, in the most common regression tasks involving neural networks, such as super-resolution \cite{image_super_resolution, enhanced_deep_residual}, segmentation \cite{segnet_a_deep, predicting_deeper_into} or video interpolation \cite{phase_based_frame, video_frame_interpolation, frame_interpolation_with}, the ratio between the input image dimension and the output image dimension also remains constant while the height and width of the output image increase. Given the previous conclusions, $\mathbf{X}_{0}$ is a rectangle of height $2m$ and width $m$. $\mathbf{X}_{1}$ is a rectangle of height $m$ and width $3m$.

\subsection{Criterion for selecting a proposed neural network} \label{subsec:3.3}
Now that the context is defined with respect to $m$, all that remains is to choose between a fully-connected neural network and a convolutional one according to $m$.

Convolutional neural networks allow better extracting 2D local image structures with fewer parameters compared to fully-connected neural networks \cite{gradient_based_learning}. This is verified by the better prediction performances of the convolutional neural networks for $m > 8$ (see Sections \ref{subsec:4.5} and \ref{subsec:4.6}). However, this property becomes less critical for small block sizes. We indeed experimentally show that, when $m \leq 8$, for the prediction problem, fully-connected architectures outperform convolutional architectures.

Therefore, when integrating the neural network based intra predictor into a codec like H.265, the criterion is to predict a block of width $m > 8$ via a proposed convolutional neural network, and a block of width $m \leq 8$ via a fully-connected neural network.

\subsection{Integration of the neural network based intra prediction into H.265} \label{subsec:3.4}
A specificity of H.265 is the quadtree structure partitioning, which determines the range of values for $m$ and the number of available decoded pixels in the context.

In H.265 \cite{high_efficiency_video}, an image is partitioned into Coding Tree Units (CTUs). A CTU is composed of one luminance Coding Tree Block (CTB), two chrominance CTBs, and syntax elements. For simplicity, let us focus on a single CTB, e.g. the luminance CTB. The CTB size is a designed parameter but the commonly used CTB size is $64 \times 64$ pixels. A CTB can be directly used as Coding Block (CB) or can be split into $4$ $32 \times 32$ CBs. Then, each $32 \times 32$ CB can be iteratively split until the size of a resulting CB reaches a minimum size. The minimum size is a designed parameter. It can be as small as $8 \times 8$ pixels, and is set to this value in most configurations. A Prediction Block (PB) is a block on which the prediction is applied. If the size of a CB is not the minimum size, this CB is identical to its PB. Otherwise, in the case of intra prediction, this CB can be split into $4$ $4 \times 4$ PBs. More splittings of this CB into PBs exist for inter prediction \cite{high_efficiency_video}. A recursive rate-distortion optimization finds the optimal splitting of each CTB.

Due to this partioning, $m \in \left\{ 4, 8, 16, 32, 64 \right\}$. For each $m \in \left\{ 4, 8 \right\}$, a fully-connected neural network is constructed with internal size $p = 1200$. Similarly, one convolutional neural network is constructed per block width $m \in \left\{ 16, 32, 64 \right\}$. The convolutional architecture for each $m \in \left\{ 16, 32, 64 \right\}$ is detailed in Appendix \ref{appendix:1}.

\begin{figure}
	\centering
	\includegraphics[width=0.48\textwidth]{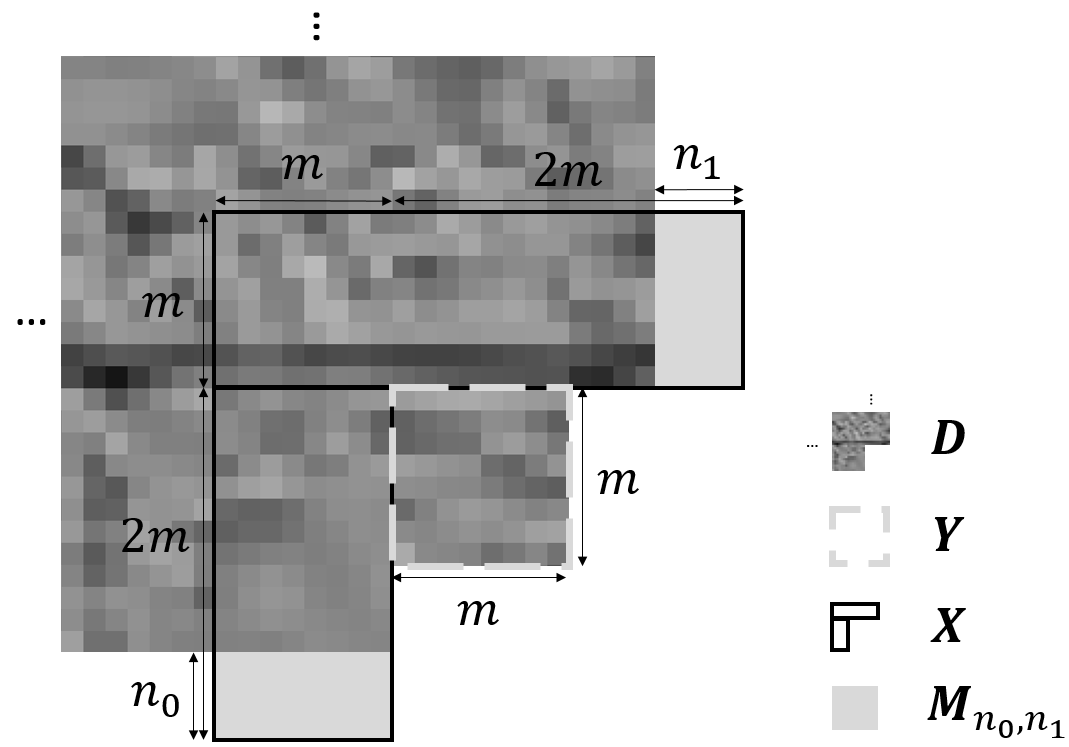}
	\caption{Illustration of the masking of the undecoded pixels in $\mathbf{X}$ for H.265. The luminance channel of the first image in the Kodak suite is being encoded via H.265 with $\text{QP} = 37$. Here, $m = 8$ and $n_{0} = n_{1} = 4$.}
	\label{fig:section3.5}
\end{figure}

Another consequence of this partitioning is that the number of available decoded pixels in the context depends on $m$ and the position of image block to be predicted in the current CTB. For instance, if the block is located at the bottom of the current CTB, the bottommost $m^{2}$ pixels in the context are not decoded yet. More generally, it might happen that a group of $n_{0} \times m$ pixels, $n_{0} \in \left\{ 0, 4, ..., m \right\}$, located at the bottom of the context, is not decoded yet. Similarly, a group of $m \times n_{1}$ pixels, $n_{1} \in \left\{ 0, 4, ..., m \right\}$, located furthest to the right in the context, may not have been decoded yet (see Figure \ref{fig:section3.5}). When pixels are not decoded yet, the solution in H.265 is to copy a decoded pixel into its neighboring undecoded pixels. But, this copy process cannot be re-used here. Indeed, it would fool the neural network and make it believe that, in an undecoded group of pixels and its surroundings, the spatial distribution of pixels intensity follows the regularity induced by the copy process. Alternatively, it is possible to indicate to a neural network that the two undecoded groups are unknown by masking them. The mask is set to the learned mean pixels intensity over the training set so that, after subtracting it from the context, the average of each input variable to a neural network over the training set is still near 0. More precisely regarding the masking, the first undecoded group in the context is fully covered by an $\alpha$-mask of height $n_{0}$ and width $m$. The second undecoded group in the context is fully covered by an $\alpha$-mask of height $m$ and width $n_{1}$ (see Figure \ref{fig:section3.5}). The two $\alpha$-masks together are denoted $\mathbf{M}_{n_{0}, n_{1}}$. In Section \ref{subsec:4.1}, the neural networks will be trained so that they adapt to this variable number of available decoded pixels in the context.

\begin{figure}
	\centering
	\includegraphics[width=0.48\textwidth]{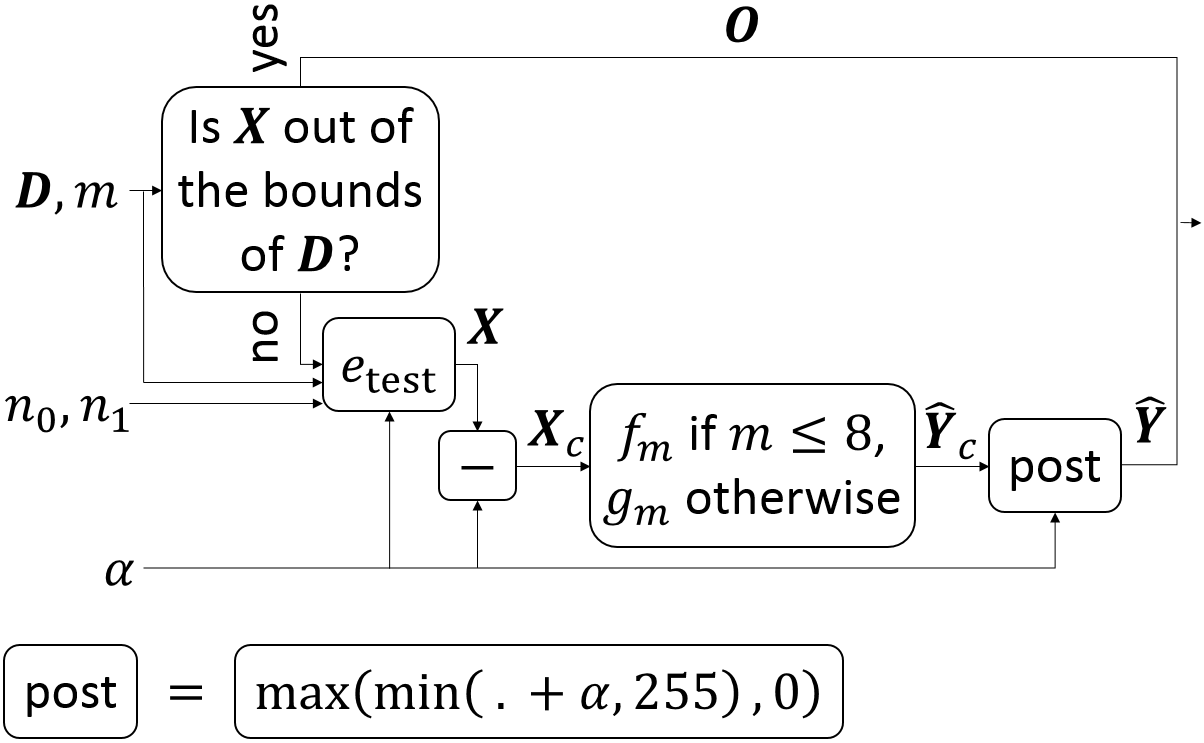}
	\caption{Illustration of the neural network based intra prediction scheme inside H.265.}
	\label{fig:section3.6}
\end{figure}

Figure \ref{fig:section3.6} summarizes the integration of the neural network based intra prediction scheme into H.265. The last issue to address is when no context is available. This occurs when the context goes out of the bounds of the decoded image, i.e. the pixel at the top-left of the context is not inside the decoded image. In this case, no neural network is used, and a zero prediction $\mathbf{O}$ of $\mathbf{Y}$ is returned. In Figure \ref{fig:section3.6}, $e_{\text{test}}$ extracts $\mathbf{X}$ from $\mathbf{D}$ with respect to the position of $\mathbf{Y}$ while applying $\mathbf{M}_{n_{0}, n_{1}}$ to $\mathbf{X}$, following Figure \ref{fig:section3.5}.


\section{Neural networks training} \label{sec:4}
This section explains how our neural networks are trained. Notably, an adaptation to the changing number of available decoded pixels in the input context is proposed. Then, an experiment shows the effectiveness of this approach. Moreover, this experiment compares the predictions of convolutional neural networks and those of fully-connected neural networks in terms of prediction PSNR in the case of large image blocks to be predicted.

\subsection{Adaptation of the neural networks to the variable number of available decoded pixels via random context masking} \label{subsec:4.1}
The fact that $n_{0}$ and $n_{1}$ vary during the test phase, e.g. in H.265, has to be considered during the training phase. It would be unpractical to train one set of neural networks for each possible pair $\left\{ n_{0}, n_{1} \right\}$. Instead, we propose to train the neural networks while feeding them with contexts containing a variable number of known pixels. More precisely, during the training phase, $n_{0}$ and $n_{1}$ are sampled uniformly from the set $\left\{ 0, 4, ..., m \right\}$. This way, the amount of available information in a training context is viewed as a random process the neural networks have to cope with.

\subsection{Objective function to be minimized} \label{subsec:4.2}
The goal of the prediction is to minimize the Euclidean distance between the image block to be predicted and its estimate, or in other words to minimize the variance of the difference between the block and its prediction \cite{source_coding_part_I}, also called the residue. The choice of the $\text{L}2$ norm is a consequence of the $\text{L}2$ norm chosen to measure the distortion between an image and its reconstruction. So, this Euclidean distance is minimized to learn the parameters $\boldsymbol{\theta}_{m}$. Moreover, regularization by $\text{L}2$ norm of the weights (not the biases), denoted $\left[ \boldsymbol{\theta}_{m} \right]_{W}$, is applied \cite{practical_recommendations_for} (see \eqref{eq:2}).
\begin{align} \label{eq:2}
	&\min_{\boldsymbol{\theta}_{m}} \; \mathbb{E} \left[ \left\Vert \mathbf{Y}_{c} - f_{m} \left( \mathbf{X}_{c}; \boldsymbol{\theta}_{m} \right) \right\Vert_{2} \right] + \lambda \left\Vert \left[ \boldsymbol{\theta}_{m} \right]_{W} \right\Vert_{2}^{2}\\
	&\mathbf{Y}_{c} = \mathbf{Y} - \alpha \nonumber
\end{align}
The expectation $\mathbb{E} \left[ . \right]$ is approximated by averaging over a training set of image blocks to be predicted, each paired with its context. For learning the parameters $\boldsymbol{\phi}_{m}$, \eqref{eq:2} is used, replacing $\boldsymbol{\theta}_{m}$ with $\boldsymbol{\phi}_{m}$ and $f_{m}$ with $g_{m}$.

The optimization algorithm is ADAM \cite{adam_a_method} with mini-batches of size $100$. The learning rate is $0.0001$ for a fully-connected neural network and $0.0004$ for a convolutional one. The number of iterations is $800000$. The learning is divided by $10$ after $400000$, $600000$, and $700000$ iterations. Regarding the other hyperparameters of ADAM, the recommended values \cite{adam_a_method} are used. The learning rates and $\lambda = 0.0005$ were found via an exhaustive search. For a given $m$, a neural network is trained for each combination of learning rate and $\lambda$ value in $\left\{ 0.007, 0.004, 0.001, 0.0007, 0.0004, 0.0001, 0.00007, 0.00004 \right\}$ $\times$ $\left\{ 0.001, 0.0005, 0.0001 \right\}$. We select the pair of parameter values giving the largest mean prediction PSNR on a validation set.

For both fully-connected and convolutional neural networks, the weights of all layers excluding the first one are initialized via Xavier's initialization \cite{understanding_the_difficulty}. Xavier's initialization allows to accelerate the training of several consecutive layers by making the variance of the input to all layers equivalent. The problem concerning the first layer is that the pixel values in the input contexts, belonging to $\left[ -\alpha, 255 - \alpha \right]$, have a large variance, which leads to instabilities during training. That is why, in the first layer, the variance of the output is reduced with respect to that of the input by initializing the weights from a Gaussian distribution with mean $0$ and standard deviation $0.01$. The biases of all layers are initialized to $0$.

\subsection{Training data} \label{subsec:4.3}
The experiments in Sections \ref{subsec:4.4} and \ref{sec:6} involve luminance images. That is why, for training, image blocks to be predicted, each paired with its context, are extracted from luminance images.

One $320 \times 320$ luminance crop $\mathbf{I}$ is, if possible, extracted from each RGB image in the ILSVRC2012 training set \cite{imagenet_a_large}. This yields a set $\mathbf{\Gamma} = \left\{ \mathbf{I}^{ \left( i \right) } \right\}_{i = 1 ... 1048717}$.

The choice of the training set of pairs of contexts and image blocks to be predicted is a critical issue. Moreover, it needs to be handled differently for fully-connected and convolutional neural networks. Indeed, a convolutional neural network predicts large image blocks with complex textures, hence its need for high power of representation (see Appendix \ref{appendix:1}). As a consequence, it overfits during training if no training data augmentation \cite{deep_big_simple, multi_column_deep, imagenet_classification_with, deep_residual_learning} is used. In constrast, a fully-connected neural network predicts small blocks with simple textures, hence its need for relatively low power of representation. Thus, it does not overfit during training without training data augmentation. Moreover, we have noticed that a training data augmentation scheme creates a bottleneck in training time for a fully-connected neural network. Therefore, training data augmentation is used for a convolutional neural network exclusively. The dependencies between a block to be predicted and its context should not be altered during the training data augmentation. Therefore, in our training data augmentation scheme, the luminance crops in $\mathbf{\Gamma}$ are exclusively randomly rotated and flipped. Precisely, for each step of ADAM, the scheme is Algorithm \ref{algo:1}. $s_{\text{rotation}}$ rotates its input image by angle $\psi \in \left\{ 0, \pi \mathbin{/} 2, \pi, 3\pi \mathbin{/} 2 \right\}$ radians. $s_{\text{flipping}}$ flips its input image horizontally with probability $0.5$. $e_{\text{train}}$ is the same function as $e_{\text{test}}$, except that $e_{\text{train}}$ extracts $\left\{ \mathbf{X}^{ \left( i \right) }, \mathbf{Y}^{ \left( i \right) } \right\}$ from potentially rotated and flipped $\mathbf{I}$ instead of extracting $\mathbf{X}$ from $\mathbf{D}$ and the position of the extraction is random instead of being defined by the order of decoding. For training a fully-connected neural network, $\left\{ \mathbf{X}_{c}^{(i)}, \mathbf{Y}_{c}^{(i)} \right\}_{i = 1 ... 10000000}$ is generated offline from $\mathbf{\Gamma}$, i.e. before the training starts.
\begin{algorithm}
	\caption{Training data augmentation for the convolutional neural networks}
	Inputs: $\mathbf{\Gamma}$, $m$, $\alpha$.
	\begin{align*}
		\forall i \in [| 1, &100 |],\\
		&\overline{i} \sim \mathcal{U} \left[ 1, 1048717 \right]\\
		&\psi \sim \mathcal{U} \left\{ 0, \frac{\pi}{2}, \pi, \frac{3\pi}{2} \right\}\\
		&n_{0}, n_{1} \sim \mathcal{U} \left\{ 0, 4, ..., m \right\}\\
		&\mathbf{J} = s_{\text{flipping}} \left( s_{\text{rotation}} \left( \mathbf{I}^{ \left( \overline{i} \right) }, \psi \right) \right)\\
		&\left\{ \mathbf{X}^{ \left( i \right) }, \mathbf{Y}^{ \left( i \right) } \right\} = e_{\text{train}} \left( \mathbf{J}, m, n_{0}, n_{1}, \alpha \right)\\
	&\mathbf{X}_{c}^{ \left( i \right) } = \mathbf{X}^{ \left( i \right) } - \alpha\\
	&\mathbf{Y}_{c}^{ \left( i \right) } = \mathbf{Y}^{ \left( i \right) } - \alpha
	\end{align*}
	Output: $\left\{ \mathbf{X}_{c}^{(i)}, \mathbf{Y}_{c}^{(i)} \right\}_{i = 1 ... 100}$.
	\label{algo:1}
\end{algorithm}

The issue regarding this generation of training data is that the training contexts have no quantization noise whereas, during the test phase in a coding scheme, a context has quantization noise. This will be discussed during several experiments in Section \ref{subsec:6.4}.

\subsection{Effectiveness of the random context masking} \label{subsec:4.4}
A natural question is whether the random context masking applied during training to adapt the neural networks to the variable number of known pixels in the context degrades the prediction performance. To address this question, a neural network trained with random context masking is compared to a set of neural networks, each trained with a fixed mask size. The experiments are performed using fully-connected neural networks for block sizes $4 \times 4$ and $8 \times 8$ pixels ($f_{4}$ and $f_{8}$), and convolutional neural networks for block sizes $16 \times 16$, $32 \times 32$, and $64 \times 64$ pixels ($g_{16}$, $g_{32}$, and $g_{64}$).

The experiments are carried out using the $24$ RGB images in the Kodak suite \cite{website_kodak}, converted into luminance. $960$ image blocks to be predicted, each paired with its context, are extracted from these luminance images. Table \ref{table:section4.1} shows the average PSNR between the image block to be predicted and its prediction via PNNS over the $960$ blocks, denoted $\text{PSNR}_{\text{PNNS}, m}$, for each block width $m$ and each test pair $\left\{ n_{0}, n_{1} \right\}$. We see that a neural network trained with a fixed mask size has performance in terms of PSNR that significantly degrades when the mask size during the training phase and the test phase differ. By contrast, a neural network trained with random context masking allows to get the best (bold) or the second best (italic) performance in terms of PSNR for all the possible mask sizes during the test phase. Moreover, when the second best PSNR performance is achieved, the second best PSNR is very close to the best one.

A second set of experiments is related to the success rate $\mu_{\text{PNNS}, m}$ of the neural network based prediction, i.e. the rate at which PNNS provides a better prediction in terms of PSNR than any other prediction of H.265. Again, the neural network trained with random context masking achieves the best (bold) or the second best rate (italic), where the second best rate is very close to the best rate (see Table \ref{table:section4.2}).

Therefore, we conclude that random context masking does not alter the performance of the trained neural networks. Besides, random context masking is an effective way of dealing with the changing number of known pixels in the context. Section \ref{sec:6} will discuss rate-distortion performance. Note that, in order to fix $n_{0}$ and $n_{1}$ during the test phase, the previous experiments have been carried out outside H.265.
\begin{table}
	\caption{Comparison of (a) $\text{PNSR}_{\text{PNNS}, 4}$, (b) $\text{PSNR}_{\text{PNNS}, 8}$, (c) $\text{PSNR}_{\text{PNNS}, 16}$, (d) $\text{PSNR}_{\text{PNNS}, 32}$, and (e) $\text{PSNR}_{\text{PNNS}, 64}$ for different pairs $\left\{ n_{0}, n_{1} \right\}$ during the training and test phases. $\mathcal{U} \left\{ 0, 4, ..., m \right\}$ denotes the fact that, during the training phase, $n_{0}$ and $n_{1}$ are uniformly drawn from the set $\left\{ 0, 4, 8, 12, ..., m \right\}$.}
	\begin{subtable}{0.48\textwidth}
		\centering
		\begin{tabular}{l|ccccc}
			\hline
			\multirow{2}{*}{Test $\left\{ n_{0}, n_{1} \right\}$} & \multicolumn{5}{c}{Training $f_{4}$ with $\left\{ n_{0}, n_{1} \right\}$}\\ \cline{2-6} & $\left\{ 0, 0 \right\}$ & $\left\{ 0, 4 \right\}$ & $\left\{ 4, 0 \right\}$ & $\left\{ 4, 4 \right\}$ & $\mathcal{U} \left\{ 0, 4 \right\} \quad $\\
			\hline
			$\left\{ 0, 0 \right\}$ & $\mathbf{34.63}$ & $34.39$ & $34.44$ & $34.23$ & $\mathit{34.57}$\\
			$\left\{ 0, 4 \right\}$ & $32.90$ & $\mathit{34.39}$ & $32.82$ & $34.23$ & $\mathbf{34.42}$\\
			$\left\{ 4, 0 \right\}$ & $32.79$ & $32.68$ & $\mathbf{34.44}$ & $34.23$ & $\mathit{34.39}$\\
			$\left\{ 4, 4 \right\}$ & $30.93$ & $32.68$ & $32.82$ & $\mathbf{34.23}$ & $\mathit{34.20}$\\
			\hline
		\end{tabular}
		\caption{}
		\label{table:section4.1.a}
	\end{subtable}
	\begin{subtable}{0.48\textwidth}
		\centering
		\begin{tabular}{l|ccccc}
			\hline
			\multirow{2}{*}{Test $\left\{ n_{0}, n_{1} \right\}$} & \multicolumn{5}{c}{Training $f_{8}$ with $\left\{ n_{0}, n_{1} \right\}$}\\ \cline{2-6} & $\left\{ 0, 0 \right\}$ & $\left\{ 0, 8 \right\}$ & $\left\{ 8, 0 \right\}$ & $\left\{ 8, 8 \right\}$ & $\mathcal{U} \left\{ 0, 4, 8 \right\} \quad $\\
			\hline
			$\left\{ 0, 0 \right\}$ & $\mathbf{31.85}$ & $31.73$ & $31.68$ & $31.62$ & $\mathit{31.79}$\\
			$\left\{ 0, 8 \right\}$ & $30.93$ & $\mathbf{31.73}$ & $30.83$ & $31.62$ & $\mathit{31.66}$\\
			$\left\{ 8, 0 \right\}$ & $30.70$ & $30.75$ & $\mathbf{31.68}$ & $31.62$ & $\mathbf{31.68}$\\
			$\left\{ 8, 8 \right\}$ & $29.58$ & $30.75$ & $30.83$ & $\mathbf{31.62}$ & $\mathit{31.56}$\\
			\hline
		\end{tabular}
		\caption{}
		\label{table:section4.1.b}
	\end{subtable}
	\begin{subtable}{0.48\textwidth}
		\centering
		\begin{tabular}{l|ccccc}
			\hline
			\multirow{2}{*}{Test $\left\{ n_{0}, n_{1} \right\}$} & \multicolumn{5}{c}{Training $g_{16}$ with $\left\{ n_{0}, n_{1} \right\}$}\\ \cline{2-6} & $\left\{ 0, 0 \right\}$ & $\left\{ 0, 16 \right\}$ & $\left\{ 16, 0 \right\}$ & $\left\{ 16, 16 \right\}$ & $\mathcal{U} \left\{ 0, 4, ..., 16 \right\}$\\
			\hline
			$\left\{ 0, 0 \right\}$ & $\mathit{29.23}$ & $22.69$ & $24.85$ & $20.76$ & $\mathbf{29.25}$\\
			$\left\{ 0, 16 \right\}$ & $28.65$ & $\mathbf{29.12}$ & $24.66$ & $23.99$ & $\mathit{29.11}$\\
			$\left\{ 16, 0 \right\}$ & $28.43$ & $22.60$ & $\mathit{29.06}$ & $22.99$ & $\mathbf{29.12}$\\
			$\left\{ 16, 16 \right\}$ & $27.87$ & $28.37$ & $28.35$ & $\mathbf{28.98}$ & $\mathit{28.97}$\\
			\hline
		\end{tabular}
		\caption{}
		\label{table:section4.1.c}
	\end{subtable}
	\begin{subtable}{0.48\textwidth}
		\centering
		\begin{tabular}{l|ccccc}
			\hline
			\multirow{2}{*}{Test $\left\{ n_{0}, n_{1} \right\}$} & \multicolumn{5}{c}{Training $g_{32}$ with $\left\{ n_{0}, n_{1} \right\}$}\\ \cline{2-6} & $\left\{ 0, 0 \right\}$ & $\left\{ 0, 32 \right\}$ & $\left\{ 32, 0 \right\}$ & $\left\{ 32, 32 \right\}$ & $\mathcal{U} \left\{ 0, 4, ..., 32 \right\}$\\
			\hline
			$\left\{ 0, 0 \right\}$ & $\mathbf{25.93}$ & $22.13$ & $22.29$ & $18.85$ & $\mathit{25.92}$\\
			$\left\{ 0, 32 \right\}$ & $25.39$ & $\mathbf{25.79}$ & $22.18$ & $21.75$ & $\mathit{25.76}$\\
			$\left\{ 32, 0 \right\}$ & $25.41$ & $22.02$ & $\mathbf{25.82}$ & $20.38$ & $\mathit{25.80}$\\
			$\left\{ 32, 32 \right\}$ & $25.00$ & $25.31$ & $25.32$ & $\mathbf{25.63}$ & $\mathit{25.62}$\\
			\hline
		\end{tabular}
		\caption{}
		\label{table:section4.1.d}
	\end{subtable}
	\begin{subtable}{0.48\textwidth}
		\centering
		\begin{tabular}{l|ccccc}
			\hline
			\multirow{2}{*}{Test $\left\{ n_{0}, n_{1} \right\}$} & \multicolumn{5}{c}{Training $g_{64}$ with $\left\{ n_{0}, n_{1} \right\}$}\\ \cline{2-6} & $\left\{ 0, 0 \right\}$ & $\left\{ 0, 64 \right\}$ & $\left\{ 64, 0 \right\}$ & $\left\{ 64, 64 \right\}$ & $\mathcal{U} \left\{ 0, 4, ..., 64 \right\}$\\
			\hline
			$\left\{ 0, 0 \right\}$ & $\mathit{21.46}$ & $19.41$ & $19.78$ & $18.17$ & $\mathbf{21.47}$\\
			$\left\{ 0, 64 \right\}$ & $21.27$ & $\mathit{21.35}$ & $19.68$ & $19.95$ & $\mathbf{21.38}$\\
			$\left\{ 64, 0 \right\}$ & $21.11$ & $19.18$ & $\mathbf{21.34}$ & $19.06$ & $\mathbf{21.34}$\\
			$\left\{ 64, 64 \right\}$ & $20.94$ & $21.08$ & $21.16$ & $\mathbf{21.27}$ & $\mathbf{21.27}$\\
			\hline
		\end{tabular}
		\caption{}
		\label{table:section4.1.e}
	\end{subtable}
	\label{table:section4.1}
\end{table}
\begin{table}
	\caption{Comparison of success rates in percentage (a) $\mu_{\text{PNNS}, 4}$, (b) $\mu_{\text{PNNS}, 8}$, (c) $\mu_{\text{PNNS}, 16}$, (d) $\mu_{\text{PNNS}, 32}$, and (e) $\mu_{\text{PNNS}, 64}$ for different pairs $\left\{ n_{0}, n_{1} \right\}$ during the training and test phases. $\mathcal{U} \left\{ 0, 4, ..., m \right\}$ denotes the fact that, during the training phase, $n_{0}$ and $n_{1}$ are uniformly drawn from the set $\left\{ 0, 4, 8, 12, ..., m \right\}$.}
	\begin{subtable}{0.48\textwidth}
		\centering
		\begin{tabular}{l|ccccc}
			\hline
			\multirow{2}{*}{Test $\left\{ n_{0}, n_{1} \right\}$} & \multicolumn{5}{c}{Training $f_{4}$ with $\left\{ n_{0}, n_{1} \right\}$}\\ \cline{2-6} & $\left\{ 0, 0 \right\}$ & $\left\{ 0, 4 \right\}$ & $\left\{ 4, 0 \right\}$ & $\left\{ 4, 4 \right\}$ & $\mathcal{U} \left\{ 0, 4 \right\}$\\
			\hline
			$\left\{ 0, 0 \right\}$ & $\mathbf{22\%}$ & $17\%$ & $19\%$ & $16\%$ & $\mathit{19\%}$\\
			$\left\{ 0, 4 \right\}$ & $15\%$ & $\mathbf{18\%}$ & $13\%$ & $15\%$ & $\mathit{17\%}$\\
			$\left\{ 4, 0 \right\}$ & $11\%$ & $11\%$ & $\mathbf{20\%}$ & $17\%$ & $\mathit{19\%}$\\
			$\left\{ 4, 4 \right\}$ & $11\%$ & $12\%$ & $13\%$ & $\mathbf{16\%}$ & $\mathit{15\%}$\\
			\hline
		\end{tabular}
		\caption{}
		\label{table:section4.2.a}
	\end{subtable}
	\begin{subtable}{0.48\textwidth}
		\centering
		\begin{tabular}{l|ccccc}
			\hline
			\multirow{2}{*}{Test $\left\{ n_{0}, n_{1} \right\}$} & \multicolumn{5}{c}{Training $f_{8}$ with $\left\{ n_{0}, n_{1} \right\}$}\\ \cline{2-6} & $\left\{ 0, 0 \right\}$ & $\left\{ 0, 8 \right\}$ & $\left\{ 8, 0 \right\}$ & $\left\{ 8, 8 \right\}$ & $\mathcal{U} \left\{ 0, 4, 8 \right\}$\\
			\hline
			$\left\{ 0, 0 \right\}$ & $\mathbf{33}\%$ & $30\%$ & $30\%$ & $30\%$ & $\mathit{32}\%$\\
			$\left\{ 0, 8 \right\}$ & $21\%$ & $\mathbf{31}\%$ & $20\%$ & $\mathbf{31}\%$ & $29\%$\\
			$\left\{ 8, 0 \right\}$ & $21\%$ & $20\%$ & $\mathbf{34}\%$ & $31\%$ & $\mathit{32}\%$\\
			$\left\{ 8, 8 \right\}$ & $16\%$ & $20\%$ & $20\%$ & $\mathbf{31}\%$ & $\mathit{30}\%$\\
			\hline
		\end{tabular}
		\caption{}
		\label{table:section4.2.b}
	\end{subtable}
	\begin{subtable}{0.48\textwidth}
		\centering
		\begin{tabular}{l|ccccc}
			\hline
			\multirow{2}{*}{Test $\left\{ n_{0}, n_{1} \right\}$} & \multicolumn{5}{c}{Training $g_{16}$ with $\left\{ n_{0}, n_{1} \right\}$}\\ \cline{2-6} & $\left\{ 0, 0 \right\}$ & $\left\{ 0, 16 \right\}$ & $\left\{ 16, 0 \right\}$ & $\left\{ 16, 16 \right\}$ & $\mathcal{U} \left\{ 0, 4, ..., 16 \right\}$\\
			\hline
			$\left\{ 0, 0 \right\}$ & $\mathbf{55\%}$ & $18\%$ & $20\%$ & $9\%$ & $\mathit{54\%}$\\
			$\left\{ 0, 16 \right\}$ & $42\%$ & $\mathbf{51\%}$ & $18\%$ & $18\%$ & $\mathit{50\%}$\\
			$\left\{ 16, 0 \right\}$ & $40\%$ & $15\%$ & $\mathit{51\%}$ & $17\%$ & $\mathbf{53\%}$\\
			$\left\{ 16, 16 \right\}$ & $33\%$ & $36\%$ & $40\%$ & $\mathbf{52\%}$ & $\mathit{49\%}$\\
			\hline
		\end{tabular}
		\caption{}
		\label{table:section4.2.c}
	\end{subtable}
	\begin{subtable}{0.48\textwidth}
		\centering
		\begin{tabular}{l|ccccc}
			\hline
			\multirow{2}{*}{Test $\left\{ n_{0}, n_{1} \right\}$} & \multicolumn{5}{c}{Training $g_{32}$ with $\left\{ n_{0}, n_{1} \right\}$}\\ \cline{2-6} & $\left\{ 0, 0 \right\}$ & $\left\{ 0, 32 \right\}$ & $\left\{ 32, 0 \right\}$ & $\left\{ 32, 32 \right\}$ & $\mathcal{U} \left\{ 0, 4, ..., 32 \right\}$\\
			\hline
			$\left\{ 0, 0 \right\}$ & $\mathbf{63\%}$ & $30\%$ & $30\%$ & $14\%$ & $\mathbf{63\%}$\\
			$\left\{ 0, 32 \right\}$ & $54\%$ & $\mathbf{61\%}$ & $27\%$ & $27\%$ & $\mathbf{61\%}$\\
			$\left\{ 32, 0 \right\}$ & $54\%$ & $28\%$ & $\mathbf{63\%}$ & $22\%$ & $\mathit{62\%}$\\
			$\left\{ 32, 32 \right\}$ & $46\%$ & $53\%$ & $52\%$ & $\mathbf{60\%}$ & $\mathbf{60\%}$\\
			\hline
		\end{tabular}
		\caption{}
		\label{table:section4.2.d}
	\end{subtable}
	\begin{subtable}{0.48\textwidth}
		\centering
		\begin{tabular}{l|ccccc}
			\hline
			\multirow{2}{*}{Test $\left\{ n_{0}, n_{1} \right\}$} & \multicolumn{5}{c}{Training $g_{64}$ with $\left\{ n_{0}, n_{1} \right\}$}\\ \cline{2-6} & $\left\{ 0, 0 \right\}$ & $\left\{ 0, 64 \right\}$ & $\left\{ 64, 0 \right\}$ & $\left\{ 64, 64 \right\}$ & $\mathcal{U} \left\{ 0, 4, ..., 64 \right\}$\\
			\hline
			$\left\{ 0, 0 \right\}$ & $\mathbf{68\%}$ & $39\%$ & $43\%$ & $27\%$ & $\mathit{67\%}$\\
			$\left\{ 0, 64 \right\}$ & $63\%$ & $\mathit{64\%}$ & $41\%$ & $44\%$ & $\mathbf{65\%}$\\
			$\left\{ 64, 0 \right\}$ & $62\%$ & $36\%$ & $\mathit{66\%}$ & $38\%$ & $\mathbf{68\%}$\\
			$\left\{ 64, 64 \right\}$ & $57\%$ & $61\%$ & $63\%$ & $\mathit{64\%}$ & $\mathbf{66\%}$\\
			\hline
		\end{tabular}
		\caption{}
		\label{table:section4.2.e}
	\end{subtable}
	\label{table:section4.2}
\end{table}

\subsection{Relevance of convolutional neural networks for predicting large image blocks} \label{subsec:4.5}
\begin{figure}
	\begin{subfigure}{0.09\textwidth}
		\centering
		\vspace{5.2mm}
		\includegraphics{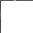}
		\caption{}
		\label{fig:section4.1.a}
	\end{subfigure}
	\begin{subfigure}{0.09\textwidth}
		\centering
		\includegraphics{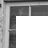}
		\caption{}
		\label{fig:section4.1.b}
	\end{subfigure}
	\begin{subfigure}{0.09\textwidth}
		\centering
		\includegraphics{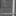}
		\caption{}
		\label{fig:section4.1.c}
	\end{subfigure}
	\begin{subfigure}{0.09\textwidth}
		\centering
		\includegraphics{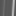}
		\caption{}
		\label{fig:section4.1.d}
	\end{subfigure}
	\begin{subfigure}{0.09\textwidth}
		\centering
		\includegraphics{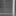}
		\caption{}
		\label{fig:section4.1.e}
	\end{subfigure}
	\caption{Prediction of a block of size $16 \times 16$ pixels via the best H.265 mode in terms of PSNR and PNNS: (a) H.265 causal neighborhood, (b) PNNS context, (c) block to be predicted, (d) predicted block via the best H.265 mode (of index $27$) in terms of PSNR, and (e) predicted block via PNNS.}
	\label{fig:section4.1}
\end{figure}

\begin{figure}
	\begin{subfigure}{0.09\textwidth}
		\centering
		\vspace{5.2mm}
		\includegraphics{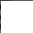}
		\caption{}
		\label{fig:section4.2.a}
	\end{subfigure}
	\begin{subfigure}{0.09\textwidth}
		\centering
		\includegraphics{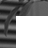}
		\caption{}
		\label{fig:section4.2.b}
	\end{subfigure}
	\begin{subfigure}{0.09\textwidth}
		\centering
		\includegraphics{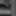}
		\caption{}
		\label{fig:section4.2.c}
	\end{subfigure}
	\begin{subfigure}{0.09\textwidth}
		\centering
		\includegraphics{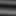}
		\caption{}
		\label{fig:section4.2.d}
	\end{subfigure}
	\begin{subfigure}{0.09\textwidth}
		\centering
		\includegraphics{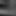}
		\caption{}
		\label{fig:section4.2.e}
	\end{subfigure}
	\caption{Prediction of a block of size $16 \times 16$ pixels via the best H.265 mode in terms of PSNR and PNNS: (a) H.265 causal neighborhood, (b) PNNS context, (c) block to be predicted, (d) predicted block via the best H.265 mode (of index $11$) in terms of PSNR, and (e) predicted block via PNNS.}
	\label{fig:section4.2}
\end{figure}

\begin{figure}
	\begin{subfigure}{0.09\textwidth}
		\centering
		\vspace{5.2mm}
		\includegraphics{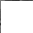}
		\caption{}
		\label{fig:section4.3.a}
	\end{subfigure}
	\begin{subfigure}{0.09\textwidth}
		\centering
		\includegraphics{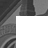}
		\caption{}
		\label{fig:section4.3.b}
	\end{subfigure}
	\begin{subfigure}{0.09\textwidth}
		\centering
		\includegraphics{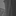}
		\caption{}
		\label{fig:section4.3.c}
	\end{subfigure}
	\begin{subfigure}{0.09\textwidth}
		\centering
		\includegraphics{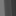}
		\caption{}
		\label{fig:section4.3.d}
	\end{subfigure}
	\begin{subfigure}{0.09\textwidth}
		\centering
		\includegraphics{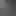}
		\caption{}
		\label{fig:section4.3.e}
	\end{subfigure}
	\caption{Prediction of a block of size $16 \times 16$ pixels via the best H.265 mode in terms of PSNR and PNNS: (a) H.265 causal neighborhood, (b) PNNS context, (c) block to be predicted, (d) predicted block via the best H.265 mode (of index $25$) in terms of PSNR, and (e) predicted block via PNNS.}
	\label{fig:section4.3}
\end{figure}

\begin{figure}
	\begin{subfigure}{0.09\textwidth}
		\centering
		\vspace{5.2mm}
		\includegraphics{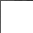}
		\caption{}
		\label{fig:section4.4.a}
	\end{subfigure}
	\begin{subfigure}{0.09\textwidth}
		\centering
		\includegraphics{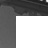}
		\caption{}
		\label{fig:section4.4.b}
	\end{subfigure}
	\begin{subfigure}{0.09\textwidth}
		\centering
		\includegraphics{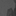}
		\caption{}
		\label{fig:section4.4.c}
	\end{subfigure}
	\begin{subfigure}{0.09\textwidth}
		\centering
		\includegraphics{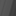}
		\caption{}
		\label{fig:section4.4.d}
	\end{subfigure}
	\begin{subfigure}{0.09\textwidth}
		\centering
		\includegraphics{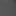}
		\caption{}
		\label{fig:section4.4.e}
	\end{subfigure}
	\caption{Prediction of a block of size $16 \times 16$ pixels via the best H.265 mode in terms of PSNR and PNNS: (a) H.265 causal neighborhood, (b) PNNS context, (c) block to be predicted, (d) predicted block via the best H.265 mode (of index $29$) in terms of PSNR, and (e) predicted block via PNNS.}
	\label{fig:section4.4}
\end{figure}

\begin{figure}
	\begin{subfigure}{0.24\textwidth}
		\centering
		\vspace{10.4mm}
		\includegraphics[scale=0.5]{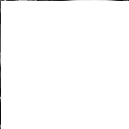}
		\caption{}
		\label{fig:section4.5.a}
	\end{subfigure}
	\begin{subfigure}{0.24\textwidth}
		\centering
		\includegraphics[scale=0.5]{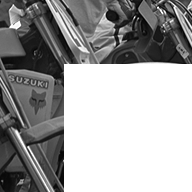}
		\caption{}
		\label{fig:section4.5.b}
	\end{subfigure}
	\begin{subfigure}{0.15\textwidth}
		\centering
		\includegraphics[scale=0.5]{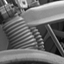}
		\caption{}
		\label{fig:section4.5.c}
	\end{subfigure}
	\begin{subfigure}{0.15\textwidth}
		\centering
		\includegraphics[scale=0.5]{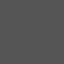}
		\caption{}
		\label{fig:section4.5.d}
	\end{subfigure}
	\begin{subfigure}{0.15\textwidth}
		\centering
		\includegraphics[scale=0.5]{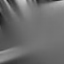}
		\caption{}
		\label{fig:section4.5.e}
	\end{subfigure}
	\caption{Prediction of a block of size $64 \times 64$ pixels via the best H.265 mode in terms of PSNR and PNNS: (a) H.265 causal neighborhood, (b) PNNS context, (c) block to be predicted, (d) predicted block via the best H.265 mode (DC) in terms of PSNR, and (e) predicted block via PNNS.}
	\label{fig:section4.5}
\end{figure}

\begin{figure}
	\begin{subfigure}{0.24\textwidth}
		\centering
		\vspace{10.4mm}
		\includegraphics[scale=0.5]{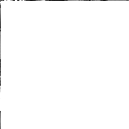}
		\caption{}
		\label{fig:section4.6.a}
	\end{subfigure}
	\begin{subfigure}{0.24\textwidth}
		\centering
		\includegraphics[scale=0.5]{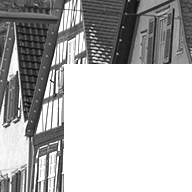}
		\caption{}
		\label{fig:section4.6.b}
	\end{subfigure}
	\begin{subfigure}{0.15\textwidth}
		\centering
		\includegraphics[scale=0.5]{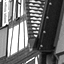}
		\caption{}
		\label{fig:section4.6.c}
	\end{subfigure}
	\begin{subfigure}{0.15\textwidth}
		\centering
		\includegraphics[scale=0.5]{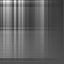}
		\caption{}
		\label{fig:section4.6.d}
	\end{subfigure}
	\begin{subfigure}{0.15\textwidth}
		\centering
		\includegraphics[scale=0.5]{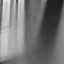}
		\caption{}
		\label{fig:section4.6.e}
	\end{subfigure}
	\caption{Prediction of a block of size $64 \times 64$ pixels via the best H.265 mode in terms of PSNR and PNNS: (a) H.265 causal neighborhood, (b) PNNS context, (c) block to be predicted, (d) predicted block via the best H.265 mode (planar) in terms of PSNR, and (e) predicted block via PNNS.}
	\label{fig:section4.6}
\end{figure}

Let us have a look at the overall efficiency of our convolutional neural networks for predicting large image blocks before comparing convolutional neural networks and fully-connected neural networks in this case. Figures \ref{fig:section4.1}, \ref{fig:section4.2}, \ref{fig:section4.3}, \ref{fig:section4.4}, \ref{fig:section4.5}, and \ref{fig:section4.6} each compares the prediction of an image block provided by the best H.265 intra prediction mode in terms of prediction PSNR and the prediction provided by PNNS. Note that, the neural networks of PNNS yielding these predictions are trained via random context masking. Note also that $n_{0} = n_{1} = 0$ during the test phase. In Figure \ref{fig:section4.5}, the image block to be predicted contains the frame of a motorbike. There is no better H.265 intra prediction mode in terms of prediction PSNR than the DC mode in this case. In contrast, PNNS can predict a coarse version of the frame of the motorbike. In Figure \ref{fig:section4.6}, the block to be predicted contains lines of various directions. PNNS predicts a combination of diagonal lines, vertical lines and horizontal lines, which is not feasible for a H.265 intra prediction mode. However, Figures \ref{fig:section4.3} and \ref{fig:section4.4} show failure cases for the convolutional neural network. The quality of the prediction provided by the convolutional neural network looks worse than that provided by the best H.265 mode. Indeed, the prediction PSNRs are $27.09$ dB and $23.85$ dB for the H.265 mode of index $25$ and the convolutional neural network respectively in Figure \ref{fig:section4.3}. They are $29.35$ dB and $27.37$ dB for the H.265 mode of index $29$ and the convolutional neural network respectively in Figure \ref{fig:section4.4}. Therefore, the convolutional neural networks of PNNS is often able to model a large set of complex textures found in large image blocks.

Table \ref{table:section4.3} compares our set of neural networks $\left\{ f_{4}, f_{8}, g_{16}, g_{32}, g_{64} \right\}$ with the set of $4$ fully-connected neural networks in \cite{fully_connected_network}, called Intra Prediction Fully-Connected Networks Single (IPFCN-S), in terms of prediction PSNR. \enquote{Single} refers to the single set of $4$ fully-connected neural networks. This differs from the two sets of fully-connected neural networks mentioned later on in Section \ref{subsec:6.3}. The $4$ fully-connected neural networks in \cite{fully_connected_network} predict image blocks of sizes $4 \times 4$, $8 \times 8$, $16 \times 16$, and $32 \times 32$ pixels respectively. Let $\text{PSNR}_{\text{IPFCN-S}, m}$ be the PSNR between the image block and its prediction via IPFCN-S, averaged over the $960$ blocks. We thank the authors in \cite{fully_connected_network} for sharing the trained model of each fully-connected neural network of IPFCN-S. Table \ref{table:section4.3} reveals that the performance of PNNS in terms of prediction PSNR is better than that of IPFCN-S for all sizes of image blocks to be predicted. More interestingly, when looking at the success rate $\mu_{\text{IPFCN-S}, m}$ of IPFCN-S, i.e. the rate at which IPFCN-S provides a better prediction in terms of PSNR than any other prediction of H.265, the difference $\mu_{\text{PNNS}, m} - \mu_{\text{IPFCN-S}, m}$ increases with $m$ (see Table \ref{table:section4.4}). This shows that convolutional neural networks are more appropriate than fully-connected ones in terms of prediction PSNR for large block sizes. Note that, in Tables \ref{table:section4.3} and \ref{table:section4.4}, there is no comparison for block width $64$ pixels as this block width is not considered in \cite{fully_connected_network}.
\begin{table}
	\caption{Comparison of $\text{PSNR}_{\text{PNNS}, m}$ and $\text{PSNR}_{\text{IPFCN-S}, m}$ \cite{fully_connected_network} for $m \in \left\{ 4, 8, 16, 32, 64 \right\}$. $f_{4}$, $f_{8}$, $g_{16}$, $g_{32}$, and $g_{64}$ are trained via random context masking. During the test phase, $n_{0} = n_{1} = 0$.}
	\centering
	\begin{tabular}{l|cc}
		\hline
		$m$ & $\text{PSNR}_{\text{PNNS}, m}$ & $\text{PSNR}_{\text{IPFCN-S}, m}$ \cite{fully_connected_network}\\
		\hline
		$4$ & $\mathbf{34.57}$ & $33.70$\\
		$8$ & $\mathbf{32.01}$ & $31.44$\\
		$16$ & $\mathbf{29.25}$ & $28.71$\\
		$32$ & $\mathbf{25.52}$ & $24.96$\\
		$64$ & $21.47$ & $-$\\
		\hline
	\end{tabular}
	\label{table:section4.3}
\end{table}
\begin{table}
	\caption{Comparison of success rates in percentage $\mu_{\text{PNNS}, m}$ and $\mu_{\text{IPFCN-S}, m}$ \cite{fully_connected_network}, for $m \in \left\{ 4, 8, 16, 32, 64 \right\}$. $f_{4}$, $f_{8}$, $g_{16}$, $g_{32}$, and $g_{64}$ are trained via random context masking. During the test phase, $n_{0} = n_{1} = 0$.}
	\centering
	\begin{tabular}{l|cc}
		\hline
		$m$ & $\mu_{\text{PNNS}, m}$ & $\mu_{\text{IPFCN-S}, m}$ \cite{fully_connected_network}\\
		\hline
		$4$ & $\mathbf{19\%}$ & $14\%$\\
		$8$ & $\mathbf{31\%}$ & $26\%$\\
		$16$ & $\mathbf{54\%}$ & $35\%$\\
		$32$ & $\mathbf{60\%}$ & $41\%$\\
		$64$ & $67\%$ & $-$\\
		\hline
	\end{tabular}
	\label{table:section4.4}
\end{table}

\subsection{Justification of fully-connected neural networks for predicting small image blocks} \label{subsec:4.6}
Now that the benefit of using convolutional neural networks rather than fully-connected ones for predicting large image blocks is shown, the choice of fully-connected neural networks for predicting small blocks must be justified. Table \ref{table:section4.5} displays the average PSNR between the image block to be predicted and its prediction over the $960$ blocks for both a fully-connected architecture and a convolutional one for each block width $m \in \left\{ 4, 8, 16 \right\}$. The two fully-connected architectures $f_{4}$ and $f_{8}$ are already presented in Figure \ref{fig:section3.2}. The fully-connected architecture $f_{16}$ is also that in Figure \ref{fig:section3.2}. The three convolutional architectures $g_{4}$, $g_{8}$, and $g_{16}$ are shown in Table \ref{table:appendix1.1} in Appendix \ref{appendix:1}. We see that the convolutional neural network provides better prediction PSNRs on average than its fully-connected counterpart only when $m > 8$. This justifies why $f_{4}$ and $f_{8}$ are selected for predicting blocks of width $4$ and $8$ pixels respectively.
\begin{table}
	\caption{Comparison of the average prediction PSNR given by the fully-connected architecture and that given by the convolutional one for different pairs $\left\{ n_{0}, n_{1} \right\}$ during the test phase: (a) $\text{PSNR}_{\text{PNNS}, 4}$ given by $f_{4}$ and $g_{4}$, (b) $\text{PSNR}_{\text{PNNS}, 8}$ given by $f_{8}$ and $g_{8}$, and (c) $\text{PSNR}_{\text{PNNS}, 16}$ given by $f_{16}$ and $g_{16}$. All neural networks are trained by drawing $n_{0}$ and $n_{1}$ uniformly from the set $\mathcal{U} \left\{ 0, 4, 8, 12, ..., m \right\}$.}
	\begin{subtable}{0.24\textwidth}
		\centering
		\begin{tabular}{l|ccccc}
			\hline
			Test $\left\{ n_{0}, n_{1} \right\}$ & $f_{4}$ & $g_{4}$\\
			\hline
			$\left\{ 0, 0 \right\}$ & $34.57$ & $33.82$\\
			$\left\{ 0, 4 \right\}$ & $34.42$ & $33.66$\\
			$\left\{ 4, 0 \right\}$ & $34.39$ & $33.79$\\
			$\left\{ 4, 4 \right\}$ & $34.20$ & $33.54$\\
			\hline
		\end{tabular}
		\caption{}
		\label{table:section4.5.a}
	\end{subtable}
	\begin{subtable}{0.24\textwidth}
		\centering
		\begin{tabular}{l|ccccc}
			\hline
			Test $\left\{ n_{0}, n_{1} \right\}$ & $f_{8}$ & $g_{8}$\\
			\hline
			$\left\{ 0, 0 \right\}$ & $31.79$ & $31.40$\\
			$\left\{ 0, 8 \right\}$ & $31.66$ & $31.41$\\
			$\left\{ 8, 0 \right\}$ & $31.68$ & $31.32$\\
			$\left\{ 8, 8 \right\}$ & $31.56$ & $31.32$\\
			\hline
		\end{tabular}
		\caption{}
		\label{table:section4.5.b}
	\end{subtable}
	\begin{subtable}{0.24\textwidth}
		\centering
		\begin{tabular}{l|ccccc}
			\hline
			Test $\left\{ n_{0}, n_{1} \right\}$ & $f_{16}$ & $g_{16}$\\
			\hline
			$\left\{ 0, 0 \right\}$ & $29.07$ & $29.25$\\
			$\left\{ 0, 16 \right\}$ & $28.95$ & $29.11$\\
			$\left\{ 16, 0 \right\}$ & $28.95$ & $29.12$\\
			$\left\{ 16, 16 \right\}$ & $28.86$ & $28.97$\\
			\hline
		\end{tabular}
		\caption{}
		\label{table:section4.5.c}
	\end{subtable}
	\label{table:section4.5}
\end{table}

\subsection{Different objective functions} \label{subsec:4.7}
In the objective function to be minimized over the neural network parameters, the distortion term involves the $L2$ norm of the difference between the image block to be predicted and its estimate (see \eqref{eq:2}). We also tried an alternative. This alternative consisted in replacing the $L2$ norm by the $L1$ norm. The choice of the $L1$ norm could be justified by the fact that, in H.265, for a given image block to be predicted, the criterion for selecting several intra prediction modes among all modes involves the sum of absolute differences between the image block to be predicted and its estimate\textsuperscript{\ref{footnote:estIntraPredLumaQT}}. But, this alternative distortion term did not yield any increase of the mean prediction PSNRs compared to those shown in Table \ref{table:section4.1}.

A common alternative to the regularization by $L2$ norm of the weights is the regularization by $L1$ norm of the weights \cite{practical_recommendations_for}. Both approaches have been tested and it was observed that the regularization by $L2$ norm of the weights slightly reduces the mean prediction error on a validation set at the end of the training.


\section{Signalling of the prediction modes in H.265} \label{sec:5}
Before moving on to the experiments in Section \ref{sec:6} where PNNS is integrated into a H.265 codec, the last issue is the signalling of the prediction modes inside H.265. Indeed, the integration of PNNS into H.265 requires to revisit the way all modes are signalled. Section \ref{sec:5} describes two ways of signalling PNNS into H.265. The purpose of setting up these two ways is to identify later on which signalling yields the largest PSNR-rate performance gains and discuss why a difference between them exists. The first signalling is the substitution of a H.265 intra prediction mode with PNNS. The second signalling is a switch between PNNS and the H.265 intra prediction modes.

\subsection{Substitution of a H.265 intra prediction mode with PNNS} \label{subsec:5.1}
Section \ref{subsec:5.1} first describes how H.265 selects the best of its $35$ intra prediction modes for predicting a given image block. Based on this, a criterion for finding the mode to be replaced with PNNS is built.

To select the best of its $35$ intra prediction modes for predicting a given image block, H.265 proceeds in two steps. During the first step, the $35$ modes compete with one another. Each mode takes the causal neighborhood of the block to compute a different prediction of the block. The sum of absolute differences between the input block and its prediction is linearly combined with the mode signalling cost, yielding the mode \enquote{fast} cost \footnote{\enquote{fast} stresses that the cost computation is relatively low.}. The modes associated to a given number of lowest \enquote{fast} costs are put into a \enquote{fast} list \footnote{See \enquote{TEncSearch::estIntraPredLumaQT} at \url{https://hevc.hhi.fraunhofer.de/trac/hevc/browser/trunk/source/Lib/TLibEncoder/TEncSearch.cpp} \label{footnote:estIntraPredLumaQT}}. During the second step, only the modes in the \enquote{fast} list compete with one another. The mode with the lowest rate-distortion cost is the best mode.

\begin{figure}
	\begin{subfigure}{0.24\textwidth}
		\centering
		\includegraphics[width=\textwidth]{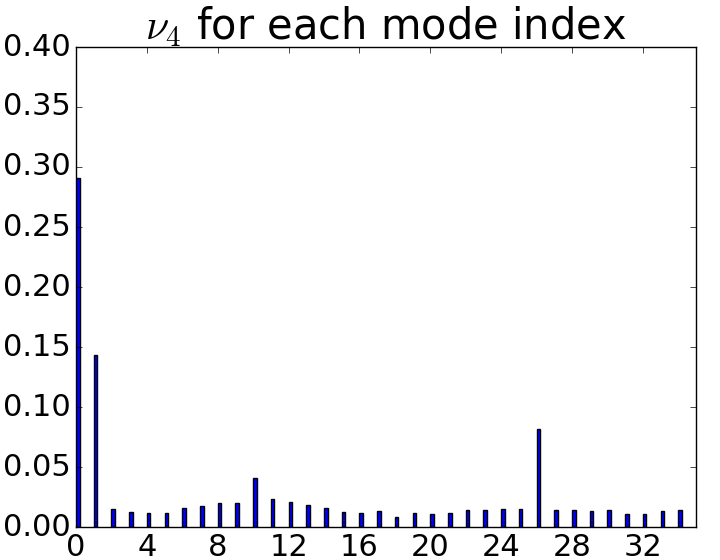}
		\caption{}
		\label{fig:section5.1.a}
	\end{subfigure}
	\begin{subfigure}{0.24\textwidth}
		\centering
		\includegraphics[width=\textwidth]{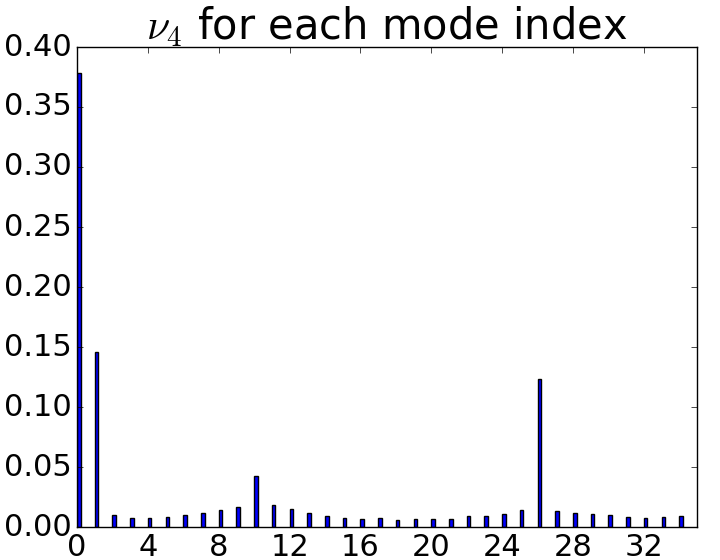}
		\caption{}
		\label{fig:section5.1.b}
	\end{subfigure}
	\begin{subfigure}{0.24\textwidth}
		\centering
		\includegraphics[width=\textwidth]{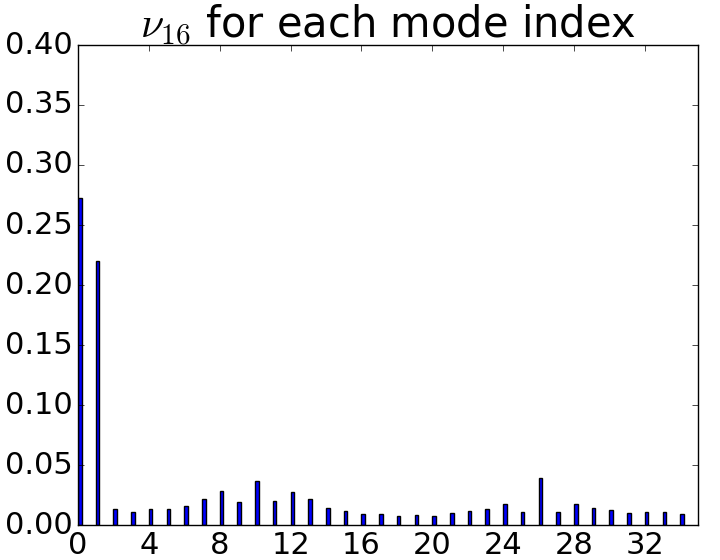}
		\caption{}
		\label{fig:section5.1.c}
	\end{subfigure}
	\begin{subfigure}{0.24\textwidth}
		\centering
		\includegraphics[width=\textwidth]{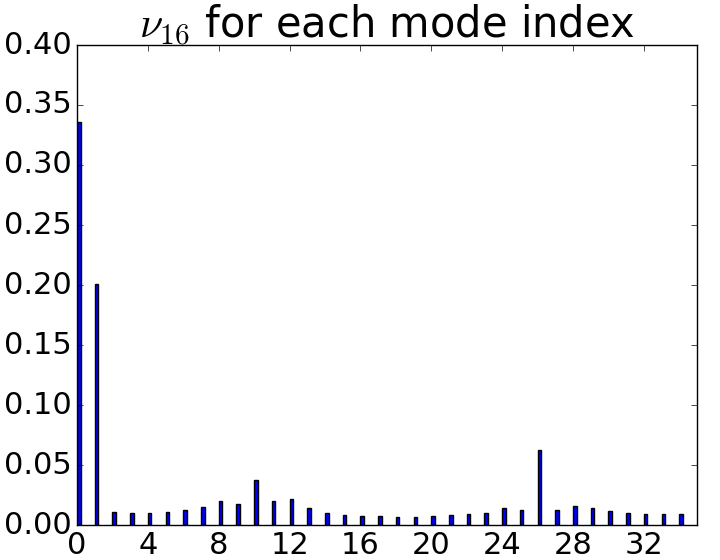}
		\caption{}
		\label{fig:section5.1.d}
	\end{subfigure}
	\begin{subfigure}{0.24\textwidth}
		\centering
		\includegraphics[width=\textwidth]{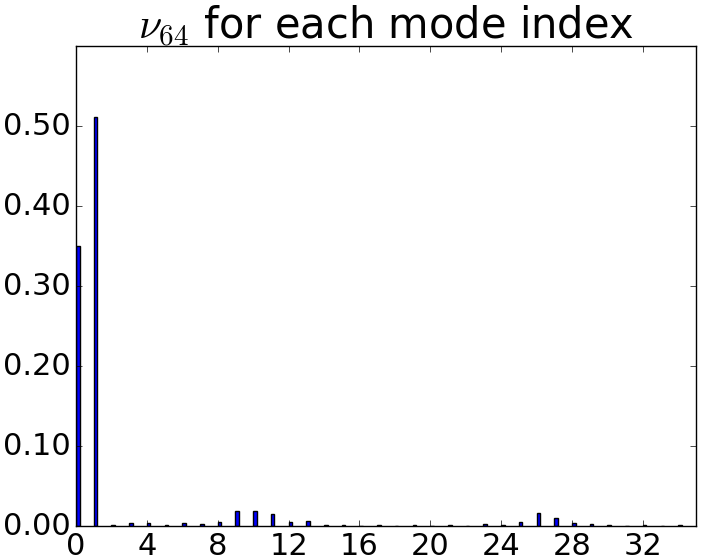}
		\caption{}
		\label{fig:section5.1.e}
	\end{subfigure}
	\begin{subfigure}{0.24\textwidth}
		\centering
		\includegraphics[width=\textwidth]{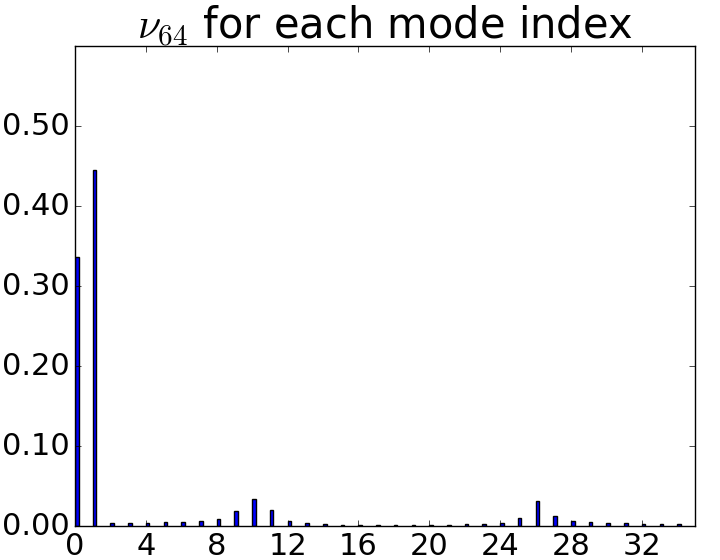}
		\caption{}
		\label{fig:section5.1.f}
	\end{subfigure}
	\caption{Analysis of (a, b) $\nu_{4}$, (c, d) $\nu_{16}$, and (e, f) $\nu_{64}$ for each mode. $100$ luminance crops from either (a, c, e) the BSDS300 dataset or (b, d, f) the INRIA Holidays dataset are encoded via H.265 with $\text{QP} = 32$.}
	\label{fig:section5.1}
\end{figure}

Knowing this, it seems natural to replace the mode that achieves the least frequently the lowest rate-distortion cost. Therefore, the frequency of interest $\nu_{\overline{m}} \in \left[ 0, 1 \right], \overline{m} \in \left\{ 4, 8, 16, 32, 64 \right\}$, is the number of times a mode has the lowest rate-distortion cost when $m = \overline{m}$ divided by the number of times the above-mentioned selection process is run when $m = \overline{m}$. To be generic, $\nu_{\overline{m}}$ should not be associated to luminance images of a specific type. That is why $100$ $380 \times 480$ luminance crops extracted from the BSDS300 dataset \cite{a_database_of} and $100$ $1200 \times 1600$ luminance crops extracted from the INRIA Holidays dataset \cite{hamming_embedding_and} are encoded with H.265 to compute $\nu_{\overline{m}}$. In this case, the mode of index $18$ has on average the lowest $\nu_{\overline{m}}$ when $\overline{m} \in \left\{ 4, 16, 32, 64 \right\}$ (see Figure \ref{fig:section5.1}). Note that this conclusion is verified with $\text{QP} \in \left\{ 22, 27, 32, 37, 42 \right\}$. Note also that statistics about the frequency of selection of each H.265 intra prediction mode have already been analyzed \cite{analysis_of_the, statistical_analysis_of}. But, they are incomplete for our case as they take into account few videos and do not differentiate each value of $\overline{m}$. Thus, PNNS replaces the H.265 mode of index $18$.

As explained thereafter, the signalling cost the H.265 intra prediction mode of index $18$ is variable and can be relatively large. When substituting the H.265 intra prediction mode of index $18$ with PNNS, this variable cost transfers to PNNS. In contrast, the purpose of the second signalling of PNNS inside H.265 is to induce a fixed and relatively low signalling cost of PNNS.

\subsection{Switch between PNNS and the H.265 intra prediction modes} \label{subsec:5.2}
The authors in \cite{fully_connected_network} propose to signal the neural network mode with a single bit. This leads to the signalling of the modes summarized in Table \ref{table:section5.1}. In addition, we modify the process of selecting the $3$ Most Probable Modes (MPMs) \cite{overview_of_the} of the current PB to make the signalling of the modes even more efficient. More precisely, if the neural network mode is the mode selected for predicting the PB above the current PB or the PB on the left side of the current PB, then the neural network mode belongs to the MPMs of the current PB. As a result, redundancy appears as the neural network mode has not only the codeword $1$ but also the codeword of a MPM. That is why, in the case where PNNS belongs to the MPMs of the current PB, we substitute each MPM being PNNS with either planar, DC or the vertical mode of index $26$ such that the $3$ MPMs of the current PB are different from each other. Besides, planar takes priority over DC, DC having priority over the vertical mode of index $26$. See the code\textsuperscript{\ref{footnote:website_code}} for further details regarding this choice.

The key remark concerning Section \ref{sec:5} is that there is a tradeoff between the signalling cost of PNNS versus the signalling cost of the H.265 intra prediction modes. Indeed, the substitution (see Section \ref{subsec:5.1}) keeps the signalling cost of the H.265 intra prediction modes constant but the signalling cost of PNNS can be up to $6$ bits. Instead, the switch decreases the signalling cost of PNNS and raises that of each H.265 intra prediction mode.
\begin{table}
	\caption{Signalling of the modes described in \cite{fully_connected_network}.}
	\centering
	\begin{tabular}{l|l}
		\hline
		Mode & codeword\\
		\hline
		Neural network mode & $1$\\
		First MPM & $010$\\
		Second MPM & $0110$\\
		Third MPM & $0111$\\
		Non-MPM and non-neural network mode & $00\left\{ 5 \text{bits} \right\}$\\
		\hline
	\end{tabular}
	\label{table:section5.1}
\end{table}


\section{Experiments} \label{sec:6}
Now that two ways of signalling PNNS inside H.265 are specified, these two ways can be compared in terms of PSNR-rate performance gains. Moreover, PNNS integrated into H.265 can be compared to IPFCN-S integrated into H.265 in terms of PSNR-rate performance gains.

\subsection{Experimental settings} \label{subsec:6.1}
The H.265 HM16.15 software is used in all the following experiments. The configuration is all-intra main. Note that the following settings only mention the integration of PNNS into H.265 via the substitution of the H.265 intra prediction mode of index $18$ with PNNS, so called \enquote{PNNS substitution}. But, the same settings apply to the integration of PNNS into H.265 via the switch between PNNS and the H.265 intra prediction modes, so called \enquote{PNNS switch}. The PSNR-rate performance of \enquote{PNNS substitution} with respect to H.265 is computed using the Bjontegaard metric \cite{calculation_of_average}, which is the average saving in bitrate of the rate-distortion curve of \enquote{PNNS substitution} with respect to the rate-distortion curve of H.265. It is interesting to analyze whether there exists a range of bitrates for which \enquote{PNNS substitution} is more beneficial. That is why $3$ different ranges of bitrates are presented. The first range, called \enquote{low rate}, refers to $\text{QP} \in \left\{ 32, 34, 37, 39, 42 \right\}$. The second range, called \enquote{high rate}, corresponds to $\text{QP} \in \left\{ 17, 19, 22, 24, 27 \right\}$. The third range, called \enquote{all rates}, computes the Bjontegaard metric with the complete set of QP values from $17$ to $42$. The H.265 common test condition \cite{common_test_conditions} recommends $\left\{ 22, 27, 32, 37 \right\}$ as QPs setting. Yet, we add several QPs to the recommended setting. This is because, to compute the Bjontegaard metric for \enquote{low rate} for instance, a polynom of degree $3$ is fitted to rate-distortion points, and at least $4$ rate-distortion points, i.e. $4$ different QPs, are required to get a good fit.

Four test sets are used to cover a wide variety of images. The first test set contains the luminance channels of respectively Barbara, Lena, Mandrill and Peppers. The second test set contains the $24$ RGB images in the Kodak suite, converted into luminance. The third test set gathers the $13$ videos sequences of the classes B, C, and D of the H.265 CTC, converted into luminance. The fourth test set contains $6$ widely used videos sequences \footnote{\url{ftp://ftp.tnt.uni-hannover.de/pub/svc/testsequences/}}, converted into luminance. Our work is dedicated to image coding. That is why only the first frame of each video sequence in the third and fourth test sets are considered. It is important to note that the training in Section \ref{sec:4}, the extraction of the frequency of selection of each H.265 intra prediction mode in Section \ref{sec:5}, and the current experiments involve $7$ distinct sets of luminance images. This way, PNNS is not tuned for any specific test luminance image.
\begin{table}
	\caption{PSNR-rate performance gains compared with H.265 of \enquote{PNNS substitution} and \enquote{PNNS switch} for the first test set.}
	\centering
	\begin{tabular}{l|cccc}
		\hline
		\multirow{3}{*}{Image name} & \multicolumn{4}{c}{PSNR-rate performance gain}\\ \cline{2-5} & \multicolumn{3}{c}{\enquote{PNNS substitution}} & \enquote{PNNS switch}\\ \cline{2-5} & Low rate & High rate & All rates & All rates\\
		\hline
		Barbara & $2.47\%$ & $1.31\%$ & $1.79\%$ & $2.77\%$\\
		Lena & $1.68\%$ & $2.11\%$ & $2.05\%$ & $3.78\%$\\
		Mandrill & $0.77\%$ & $0.58\%$ & $0.67\%$ & $1.46\%$\\
		Peppers & $1.61\%$ & $1.50\%$ & $1.71\%$ & $3.63\%$\\
		\hline
	\end{tabular}
	\label{table:section6.1}
\end{table}
\begin{table}
	\caption{PSNR-rate performance gains compared with H.265 of \enquote{PNNS substitution} and \enquote{PNNS switch} for the second test set.}
	\centering
	\begin{tabular}{l|cccc}
		\hline
		\multirow{3}{*}{Kodak image index} & \multicolumn{4}{c}{PSNR-rate performance gain}\\ \cline{2-5} & \multicolumn{3}{c}{\enquote{PNNS substitution}} & \enquote{PNNS switch}\\ \cline{2-5} & Low rate & High rate & All rates & All rates\\
		\hline
		$1$ & $1.20\%$ & $0.74\%$ & $0.95\%$ & $2.06\%$\\
		$2$ & $0.59\%$ & $0.91\%$ & $0.88\%$ & $2.16\%$\\
		$3$ & $0.91\%$ & $2.04\%$ & $1.59\%$ & $3.22\%$\\
		$4$ & $1.78\%$ & $1.75\%$ & $1.80\%$ & $3.23\%$\\
		$5$ & $1.40\%$ & $2.45\%$ & $2.08\%$ & $4.01\%$\\
		$6$ & $1.43\%$ & $0.81\%$ & $1.12\%$ & $2.11\%$\\
		$7$ & $1.12\%$ & $2.36\%$ & $1.76\%$ & $3.86\%$\\
		$8$ & $1.01\%$ & $0.83\%$ & $0.98\%$ & $1.79\%$\\
		$9$ & $1.54\%$ & $1.43\%$ & $1.46\%$ & $3.05\%$\\
		$10$ & $2.20\%$ & $2.37\%$ & $2.42\%$ & $3.84\%$\\
		$11$ & $0.93\%$ & $0.91\%$ & $1.00\%$ & $2.41\%$\\
		$12$ & $0.96\%$ & $1.02\%$ & $1.07\%$ & $2.33\%$\\
		$13$ & $0.83\%$ & $0.5\%$ & $0.64\%$ & $1.76\%$\\
		$14$ & $0.96\%$ & $1.28\%$ & $1.20\%$ & $2.76\%$\\
		$15$ & $1.53\%$ & $1.19\%$ & $1.37\%$ & $2.62\%$\\
		$16$ & $0.66\%$ & $0.62\%$ & $0.70\%$ & $1.69\%$\\
		$17$ & $1.35\%$ & $2.03\%$ & $1.80\%$ & $3.69\%$\\
		$18$ & $0.68\%$ & $0.96\%$ & $0.88\%$ & $1.98\%$\\
		$19$ & $1.44\%$ & $0.86\%$ & $1.05\%$ & $2.06\%$\\
		$20$ & $0.92\%$ & $1.61\%$ & $1.38\%$ & $2.71\%$\\
		$21$ & $0.99\%$ & $0.83\%$ & $0.94\%$ & $2.28\%$\\
		$22$ & $0.56\%$ & $0.88\%$ & $0.78\%$ & $2.22\%$\\
		$23$ & $1.20\%$ & $2.45\%$ & $2.03\%$ & $4.20\%$\\
		$24$ & $0.68\%$ & $0.87\%$ & $0.80\%$ & $1.73\%$\\
		\hline
	\end{tabular}
	\label{table:section6.2}
\end{table}
\begin{table}
	\caption{PSNR-rate performance gains compared with H.265 of \enquote{PNNS substitution} and \enquote{PNNS switch} for the third test set.}
	\centering
	\begin{tabular}{l|l|cccc}
		\hline
		\multicolumn{2}{c|}{\multirow{3}{*}{Video sequence}} & \multicolumn{4}{c}{PSNR-rate performance gain}\\ \cline{3-6} \multicolumn{2}{c|}{} & \multicolumn{3}{c}{\enquote{PNNS substitution}} & \enquote{PNNS switch}\\ \cline{3-6} \multicolumn{2}{c|}{} & Low rate & High rate & All rates & All rates\\
		\hline
		\multirow{5}{*}{B} & BQTerrace & $1.66\%$ & $0.95\%$ & $1.29\%$ & $2.44\%$\\ & BasketballDrive & $4.80\%$ & $2.87\%$ & $3.65\%$ & $5.20\%$\\ & Cactus & $1.48\%$ & $1.51\%$ & $1.58\%$ & $3.05\%$\\ & ParkScene & $0.64\%$ & $1.16\%$ & $0.97\%$ & $2.58\%$\\ & Kimono & $1.28\%$ & $1.55\%$ & $1.65\%$ & $2.92\%$\\
		\hline
		\multirow{4}{*}{C} & BQMall& $1.20\%$ & $1.30\%$ & $1.30\%$ & $3.14\%$\\ & BasketballDrill & $-1.18\%$ & $1.34\%$ & $0.39\%$ & $3.50\%$\\ & RaceHorsesC & $1.34\%$ & $1.58\%$ & $1.53\%$ & $3.29\%$\\ & PartyScene & $1.02\%$ & $0.91\%$ & $0.96\%$ & $2.42\%$\\
		\hline
		\multirow{4}{*}{D} & BQSquare & $0.79\%$ & $0.86\%$ & $0.86\%$ & $2.21\%$\\ & BasketballPass & $1.61\%$ & $1.80\%$ & $1.48\%$ & $3.08\%$\\ & BlowingBubbles & $0.66\%$ & $1.22\%$ & $1.02\%$ & $2.65\%$\\ & RaceHorses & $1.32\%$ & $1.63\%$ & $1.54\%$ & $3.28\%$\\
		\hline
	\end{tabular}
	\label{table:section6.3}
\end{table}
\begin{table}
	\caption{PSNR-rate performance gains compared with H.265 of \enquote{PNNS substitution} and \enquote{PNNS switch} for the fourth test set.}
	\centering
	\begin{tabular}{l|cccc}
		\hline
		\multirow{3}{*}{Video sequence} & \multicolumn{4}{c}{PSNR-rate performance gain}\\ \cline{2-5} & \multicolumn{3}{c}{\enquote{PNNS substitution}} & \enquote{PNNS switch}\\ \cline{2-5} & Low rate & High rate & All rates & All rates\\
		\hline
		Bus & $1.67\%$ & $1.17\%$ & $1.45\%$ & $2.74\%$\\
		City & $1.55\%$ & $1.19\%$ & $1.35\%$ & $2.51\%$\\
		Crew & $1.56\%$ & $1.24\%$ & $1.38\%$ & $3.10\%$\\
		Football & $1.44\%$ & $1.78\%$ & $1.78\%$ & $3.52\%$\\
		Harbour & $1.80\%$ & $0.73\%$ & $1.25\%$ & $2.51\%$\\
		Soccer & $0.96\%$ & $0.95\%$ & $1.03\%$ & $1.90\%$\\
		\hline
	\end{tabular}
	\label{table:section6.4}
\end{table}

\subsection{Analysis of the two ways of signalling the PNNS mode inside H.265} \label{subsec:6.2}
The most striking observation is that the PSNR-rate performance gains generated by \enquote{PNNS switch} are always larger than those provided by \enquote{PNNS substitution} (see Tables \ref{table:section6.1}, \ref{table:section6.2}, \ref{table:section6.3}, and \ref{table:section6.4}). This has two explanations. Firstly, \enquote{PNNS substitution} is hampered by the suppression of the original H.265 intra prediction mode of index $18$. Indeed, the PSNR-rate performance gain is degraded when the original H.265 intra prediction mode of index $18$ is a relevant mode for encoding a luminance image. The most telling example is the luminance channel of the first frame of \enquote{BasketballDrill}. When this channel is encoded via H.265, for the original H.265 intra prediction mode of index $18$, $\nu_{4} = 0.085$, $\nu_{8} = 0.100$, $\nu_{16} = 0.116$, $\nu_{32} = 0.085$, and $\nu_{64} = 0.088$. This means that, compared to the average statistics in Figure \ref{fig:section5.1}, the original H.265 intra prediction mode of index $18$ is used approximatively $10$ times more frequently. This explains why the PSNR-rate performance gain provided by \enquote{PNNS substitution} is only $0.39\%$ (see Table \ref{table:section6.3}). The other way round, the luminance channel of the first frame of \enquote{BasketballDrive} is an insightful example. When this channel is encoded via H.265, for the original H.265 intra prediction mode of index $18$, $\nu_{4} = 0.004$, $\nu_{8} = 0.005$, $\nu_{16} = 0.004$, $\nu_{32} = 0.004$, and $\nu_{64} = 0.000$. In this case, the original H.265 intra prediction mode of index $18$ is almost never used. \enquote{PNNS substitution} thus yields $3.65\%$ of PSNR-rate performance gain.

\begin{figure*}
	\begin{subfigure}{0.32\textwidth}
		\centering
		\includegraphics{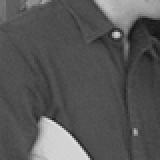}
		\caption{}
		\label{fig:section6.1.a}
	\end{subfigure}
	\begin{subfigure}{0.32\textwidth}
		\centering
		\includegraphics{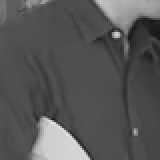}
		\caption{}
		\label{fig:section6.1.b}
	\end{subfigure}
	\begin{subfigure}{0.32\textwidth}
		\centering
		\includegraphics{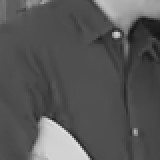}
		\caption{}
		\label{fig:section6.1.c}
	\end{subfigure}
	\caption{Comparison of (a) a $100 \times 100$ crop of the luminance channel of the first frame of BQMall, (b) its reconstruction via H.265, and (c) its reconstruction via \enquote{PNNS switch}. $\text{QP} = 27$. For the luminance channel of the first frame of BQMall, for H.265, $\left\{ \text{rate} = 0.670 \; \text{bpp}, \text{PSNR} = 38.569 \; \text{dB} \right\}$. For \enquote{PNNS switch}, $\left\{ \text{rate} = 0.644 \; \text{bpp}, \text{PSNR} = 38.513 \; \text{dB} \right\}$.}
	\label{fig:section6.1}
\end{figure*}

\begin{figure*}
	\begin{subfigure}{0.32\textwidth}
		\centering
		\includegraphics{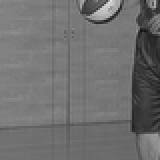}
		\caption{}
		\label{fig:section6.2.a}
	\end{subfigure}
	\begin{subfigure}{0.32\textwidth}
		\centering
		\includegraphics{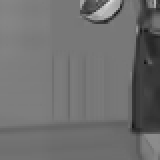}
		\caption{}
		\label{fig:section6.2.b}
	\end{subfigure}
	\begin{subfigure}{0.32\textwidth}
		\centering
		\includegraphics{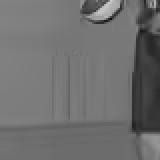}
		\caption{}
		\label{fig:section6.2.c}
	\end{subfigure}
	\caption{Comparison of (a) a $100 \times 100$ crop of the luminance channel of the first frame of BasketballPass, (b) its reconstruction via H.265, and (c) its reconstruction via \enquote{PNNS switch}. $\text{QP} = 37$. For the luminance channel of the first frame of BasketballPass, for H.265, $\left\{ \text{rate} = 0.167 \; \text{bpp}, \text{PSNR} = 32.386 \; \text{dB} \right\}$. For \enquote{PNNS switch}, $\left\{ \text{rate} = 0.160 \; \text{bpp}, \text{PSNR} = 32.407 \; \text{dB} \right\}$.}
	\label{fig:section6.2}
\end{figure*}

There is another explanation for the gap in PSNR-rate performance gain between \enquote{PNNS substitution} and \enquote{PNNS switch}. As shown in Section \ref{subsec:4.5}, PNNS is able to model a large set of complex textures found in large image blocks. PNNS is also able to model a large set of simple textures found in small blocks. Following the principle of Huffman Coding, an intra prediction mode that gives on average predictions of good quality, such as PNNS, should be signalled using fewer bits. However, an intra prediction mode that seldom yields the highest prediction quality, such as the H.265 intra prediction mode of index $4$, should be signalled using more bits. This corresponds exactly to the principle of the switch between PNNS and the H.265 intra prediction modes. Therefore, \enquote{PNNS switch} beats \enquote{PNNS substitution} in terms of PSNR-rate performance gains. Figures \ref{fig:section6.1} and \ref{fig:section6.2} each compares the reconstruction of a luminance image via H.265 and its reconstruction via \enquote{PNNS switch} at similar reconstruction PSNRs. More visual comparisons are available on the website\footnote{\url{https://www.irisa.fr/temics/demos/prediction_neural_network/PredictionNeuralNetwork.htm} \label{footnote:website_visualizations}}.

Another interesting conclusion emerges when comparing \enquote{low rate} and \enquote{high rate}. There is no specific range of bitrate for which \enquote{PNNS substitution} is more profitable (see Tables \ref{table:section6.1}, \ref{table:section6.2}, \ref{table:section6.3}, and \ref{table:section6.4}). Note that, in few cases, the PNSR-rate performance gain in \enquote{all rates} is slightly larger than those in \enquote{low rate} and \enquote{high rate}. This happens when the area between the rate-distortion curve of \enquote{PNNS substitution} and the rate-distortion curve of H.265 gets relatively large in the range $\text{QP} \in \left[ 27, 32 \right]$.

\subsection{Comparison with the state-of-the-art} \label{subsec:6.3}
Now, \enquote{PNNS switch} is compared to IPFCN-S integrated into H.265 in terms of PSNR-rate performance gains. It is important to note that the authors in \cite{fully_connected_network} develop two versions of their set of $4$ fully-connected neural networks for intra prediction. The first version, called IPFCN-S, is the one used in Section \ref{subsec:4.5}. The $4$ fully-connected neural networks are trained on an unconstrained training set of image blocks to be predicted, each paired with its context. The second version is called Intra Prediction Fully-Connected Networks Double (IPFCN-D). The training data are dissociated into two groups. One group gathers image blocks exhibiting textures with angular directions, each paired with its context. The other group gathers image blocks exhibiting textures with non-angular directions, each paired with its context. In IPFCN-D, there are two sets of $4$ fully-connected neural networks, each set being trained on a different group of training data. Then, the two sets are integrated into H.265. IPFCN-D gives slightly larger PSNR-rate performance gains than IPFCN-S. The comparison below involves IPFCN-S as our training set is not dissociated. But, this dissociation could also be applied to the training set of the neural networks of PNNS.

\enquote{PNNS switch} and IPFCN-S integrated into H.265 are compared on the third test set. The PSNR-rate performance gains of IPFCN-S are reported from \cite{fully_connected_network}. We observe that the PSNR-rate performance gains of \enquote{PNNS switch} are larger than those of IPFCN-S integrated into H.265, apart from the case of the video sequence \enquote{ParkScene} (see Table \ref{table:section6.5}). Note that, for several videos sequences, the difference in PSNR-rate performance gains between \enquote{PNNS switch} and IPFCN-S integrated into H.265 is significant. For instance, for the video sequence \enquote{BasketballPass}, the PSNR-rate performance gain of \enquote{PNNS switch} is $3.08\%$ whereas that of IPFCN-S integrated into H.265 is $1.1\%$. Therefore, the use of both fully-connected neural networks and convolutional neural networks for intra prediction, the training with random context masking and the training data augmentation for training the convolutional neural networks of PNNS help boost the PSNR-rate performance gains. This is consistent with the conclusion in Section \ref{subsec:4.5}. Note that, even when comparing the PSNR-rate performance gains of \enquote{PNNS switch} with those of IPFCN-D integrated into H.265 which are reported in \cite{fully_connected_network}, \enquote{PNNS switch} often yields larger gains.
\begin{table}
	\caption{PSNR-rate performance gains  of our proposed \enquote{PNNS switch} and IPFCN-S \cite{fully_connected_network} inside H.265 for the third test set. The reference is H.265.}
	\centering
	\begin{tabular}{l|l|cc}
		\hline
		\multicolumn{2}{c|}{\multirow{2}{*}{Video sequence}} & \multicolumn{2}{c}{PSNR-rate performance gain}\\ \cline{3-4} \multicolumn{2}{c|}{} & our \enquote{PNNS switch} & IPFCN-S \cite{fully_connected_network} inside H.265\\
		\hline
		\multirow{5}{*}{B} & BQTerrace & $\mathbf{2.44\%}$ & $1.8\%$\\ & BasketballDrive & $\mathbf{5.20\%}$ & $3.4\%$\\ & Cactus & $\mathbf{3.05\%}$ & $2.7\%$\\ & ParkScene & $2.58\%$ & $\mathbf{2.8\%}$\\ & Kimono & $\mathbf{2.92\%}$ & $2.7\%$\\
		\hline
		\multirow{4}{*}{C} & BQMall& $\mathbf{3.14\%}$ & $2.0\%$\\ & BasketballDrill & $\mathbf{3.50\%}$ & $1.1\%$\\ & RaceHorsesC & $\mathbf{3.29\%}$ & $2.9\%$\\ & PartyScene & $\mathbf{2.42\%}$ & $1.3\%$\\
		\hline
		\multirow{4}{*}{D} & BQSquare & $\mathbf{2.21\%}$ & $0.6\%$\\ & BasketballPass & $\mathbf{3.08\%}$ & $1.1\%$\\ & BlowingBubbles & $\mathbf{2.65\%}$ & $1.6\%$\\ & RaceHorses & $\mathbf{3.28\%}$ & $2.8\%$\\
		\hline
	\end{tabular}
	\label{table:section6.5}
\end{table}

\subsection{Robustness of the neural networks to quantization noise in their input context} \label{subsec:6.4}
In this section, we show that the proposed PNNS performs well with test data compressed with different QP values despite the fact that the neural networks of PNNS have been trained only once with no quantization noise in their input context. Indeed, let us consider two different \enquote{PNNS switch}. In the first \enquote{PNNS switch}, our $5$ neural networks, one for each block size, are dedicated to all QPs. Note that the first \enquote{PNNS switch} corresponds to the \enquote{PNNS switch} that has been used so far. In the second \enquote{PNNS switch}, a first set of $5$ neural networks is dedicated to $\text{QP} \leq 27$ whereas a second set is dedicated to $\text{QP} > 27$. Unlike the first set of neural networks, the second set is trained on contexts that are encoded and decoded via H.265 with $\text{QP} \sim \mathcal{U} \left\{ 32, 37, 42 \right\}$ for each training context. For the third test set, the difference in PSNR-rate performance gain between the first \enquote{PNNS switch} and the second \enquote{PNNS switch} ranges between $0.0\%$ and $0.1\%$. This means that there is no need to train the neural networks of PNNS on contexts with quantization noise.

\subsection{Complexity} \label{subsec:6.5}
A fully-connected neural network needs an overcomplete representation to provide predictions with high quality. That is why the number of neurons in each fully-connected layer is usually much larger than the size of the context. Likewise, the number of feature maps in each convolutional layer of a convolutional neural network is usually large. This incurs a high computational cost. Table \ref{table:section6.6} gives the encoding and decoding times for \enquote{PNNS switch} and IPFCN-S and shows comparable running times for both solutions. A Bi-Xeon CPU E5-2620 is used for \enquote{PNNS switch}.
\begin{table}
	\caption{Average computation time ratio with respect to H.265.}
	\centering
	\begin{tabular}{l|ccc}
		\hline
		\quad & our \enquote{PNNS switch} & IPFCN-S \cite{fully_connected_network} inside H.265\\
		\hline
		Encoding & $51$ & $46$\\
		Decoding & $191$ & $190$\\
		\hline
	\end{tabular}
	\label{table:section6.6}
\end{table}

\subsection{Memory consumption} \label{subsec:6.6}
In addition to the complexity, another issue arising from the integration of PNNS into H.265 is the increase of the memory consumption. For a regular intra prediction mode in H.265, the context around the image block to be predicted consists of one row of $2m + 1$ pixels above the block and one column of $2m$ pixels on the left side of the block. However, for a neural network of PNNS, the context around the image block consists of $m$ rows of $3m$ pixels above the block and $m$ columns of $2m$ pixels on the left side of the block (see Figure \ref{fig:section3.5}). Therefore, the hardware must be adapted to handle the large memory required by the input contexts to the neural networks of PNNS.


\section{Conclusion} \label{sec:7}
This paper has presented a set of neural network architectures, including both fully-connected neural networks and convolutional neural networks, for intra prediction. It is shown that fully-connected neural networks are well adapted to the prediction of image blocks of small sizes whereas convolutional ones provide better predictions in large blocks. Our neural networks are trained via a random context masking of their context so that they adapt to the variable number of available decoded pixels for the prediction in a coding scheme. When integrated into a H.265 codec, the proposed neural networks are shown to give rate-distortion performance gains compared with the H.265 intra prediction. Moreover, it is shown that these neural networks can cope with the quantization noise present in the prediction context, i.e. they can be trained on undistorted contexts, and then generalize well on distorted contexts in a coding scheme. This greatly simplifies training as quantization noise does not need to be taken into account during training.

\appendices

\section{Convolutional architectures for H.265} \label{appendix:1}
The architecture of the stack of convolutional layers $g_{m}^{c} \left( \; . \; ; \boldsymbol{\phi}_{m}^{c, 0} \right)$ that is applied to the context portion $\mathbf{X}_{0}$ located on the left side of the image block to be predicted is shown in Table \ref{table:appendix1.1} for $m \in \left\{ 4, 16, 64 \right\}$. The architecture of the stack of convolutional layers $g_{m}^{c} \left( \; . \; ; \boldsymbol{\phi}_{m}^{c, 1} \right)$ that is applied to the context porton $\mathbf{X}_{1}$ located above the image block is identical to that of $g_{m}^{c} \left( \; . \; ; \boldsymbol{\phi}_{m}^{c, 0} \right)$. The slope of LeakyReLU is equal to $0.1$.

The architecture of $g_{8}^{c} \left( \; . \; ; \boldsymbol{\phi}_{8}^{c, 0} \right)$ is similar to that of $g_{4}^{c} \left( \; . \; ; \boldsymbol{\phi}_{4}^{c, 0} \right)$ but, in the first layer, the filter size is $5 \times 5 \times 1$, the number of filters is $64$, and the stride is $2$. Moreover, in the second layer, the filter size is $3 \times 3 \times 64$ and the number of filters is $64$. The architecture of $g_{32}^{c} \left( \; . \; ; \boldsymbol{\phi}_{32}^{c, 0} \right)$ is similar to that of $g_{64}^{c} \left( \; . \; ; \boldsymbol{\phi}_{64}^{c, 0} \right)$ but, in the third layer, the filter size is $3 \times 3 \times 128$, the number of filters is $128$, and the stride is $1$. Moreover, in the fourth layer, the filter size is $5 \times 5 \times 128$ and the number of filters is $256$. In the fifth layer, the filter size is $3 \times 3 \times 256$ and the number of filters is $256$.

To obtain the architecture of $g_{m}^{t}$, each sequence of layers in $g_{m}^{c}$ is reversed. Besides, each convolution is replaced by a transposed convolution (see Table \ref{table:appendix1.1}).
\begin{table}
	\caption{Architectures of the stack of convolutional layers $g_{m}^{c} \left( \; . \; ; \boldsymbol{\phi}_{m}^{c, 0} \right)$ that is applied to the context portion $\mathbf{X}_{0}$ located on the left side of the image block to be predicted ((a) $m = 4$, (c) $m = 16$, and (e) $m = 64$) and the stack of transposed convolutional layers $g_{m}^{t} \left( \; . \; ; \boldsymbol{\phi}_{m}^{t} \right)$ ((b) $m = 4$, (d) $m = 16$, and (f) $m = 64$). \enquote{conv} means convolution and \enquote{tconv} means transposed convolution.}
	\begin{subtable}{0.48\textwidth}
		\centering
		\begin{tabular}{|c|c|c|c|c|c|}
			\hline
			Layer & Layer & \multirow{2}{*}{Filter size} & Number of & \multirow{2}{*}{Stride} & \multirow{2}{*}{Non-linearity}\\
			number & type & & filters & &\\
			\hline
			$1$ & conv & $3 \times 3 \times 1$ & $32$ & $1$ & LeakyReLU\\
			$2$ & conv & $3 \times 3 \times 32$ & $32$ & $1$ & LeakyReLU\\
			\hline
		\end{tabular}
		\caption{}
		\label{table:appendix1.1.a}
	\end{subtable}
	\begin{subtable}{0.48\textwidth}
		\centering
		\begin{tabular}{|c|c|c|c|c|c|}
			\hline
			Layer & Layer & \multirow{2}{*}{Filter size} & Number of & \multirow{2}{*}{Stride} & \multirow{2}{*}{Non-linearity}\\
			number & type & & filters & &\\
			\hline
			$1$ & tconv & $3 \times 3 \times 32$ & $32$ & $1$ & LeakyReLU\\
			$2$ & tconv & $3 \times 3 \times 32$ & $1$ & $1$ & -\\
			\hline
		\end{tabular}
		\caption{}
		\label{table:appendix1.1.b}
	\end{subtable}
	\begin{subtable}{0.48\textwidth}
		\centering
		\begin{tabular}{|c|c|c|c|c|c|}
			\hline
			Layer & Layer & \multirow{2}{*}{Filter size} & Number of & \multirow{2}{*}{Stride} & \multirow{2}{*}{Non-linearity}\\
			number & type & & filters & &\\
			\hline
			$1$ & conv & $5 \times 5 \times 1$ & $64$ & $2$ & LeakyReLU\\
			$2$ & conv & $3 \times 3 \times 64$ & $64$ & $1$ & LeakyReLU\\
			$3$ & conv & $5 \times 5 \times 64$ & $128$ & $2$ & LeakyReLU\\
			$4$ & conv & $3 \times 3 \times 128$ & $128$ & $1$ & LeakyReLU\\
			\hline
		\end{tabular}
		\caption{}
		\label{table:appendix1.1.c}
	\end{subtable}
	\begin{subtable}{0.48\textwidth}
		\centering
		\begin{tabular}{|c|c|c|c|c|c|}
			\hline
			Layer & Layer & \multirow{2}{*}{Filter size} & Number of & \multirow{2}{*}{Stride} & \multirow{2}{*}{Non-linearity}\\
			number & type & & filters & &\\
			\hline
			$1$ & tconv & $3 \times 3 \times 128$ & $128$ & $1$ & LeakyReLU\\
			$2$ & tconv & $5 \times 5 \times 128$ & $64$ & $2$ & LeakyReLU\\
			$3$ & tconv & $3 \times 3 \times 64$ & $64$ & $1$ & LeakyReLU\\
			$4$ & tconv & $5 \times 5 \times 64$ & $1$ & $2$ & -\\
			\hline
		\end{tabular}
		\caption{}
		\label{table:appendix1.1.d}
	\end{subtable}
	\begin{subtable}{0.48\textwidth}
		\centering
		\begin{tabular}{|c|c|c|c|c|c|}
			\hline
			Layer & Layer & \multirow{2}{*}{Filter size} & Number of & \multirow{2}{*}{Stride} & \multirow{2}{*}{Non-linearity}\\
			number & type & & filters & &\\
			\hline
			$1$ & conv & $5 \times 5 \times 1$ & $64$ & $2$ & LeakyReLU\\
			$2$ & conv & $5 \times 5 \times 64$ & $128$ & $2$ & LeakyReLU\\
			$3$ & conv & $5 \times 5 \times 128$ & $256$ & $2$ & LeakyReLU\\
			$4$ & conv & $5 \times 5 \times 256$ & $512$ & $2$ & LeakyReLU\\
			$5$ & conv & $3 \times 3 \times 512$ & $512$ & $1$ & LeakyReLU\\
			\hline
		\end{tabular}
		\caption{}
		\label{table:appendix1.1.e}
	\end{subtable}
	\begin{subtable}{0.48\textwidth}
		\centering
		\begin{tabular}{|c|c|c|c|c|c|}
			\hline
			Layer & Layer & \multirow{2}{*}{Filter size} & Number of & \multirow{2}{*}{Stride} & \multirow{2}{*}{Non-linearity}\\
			number & type & & filters & &\\
			\hline
			$1$ & tconv & $3 \times 3 \times 512$ & $512$ & $1$ & LeakyReLU\\
			$2$ & tconv & $5 \times 5 \times 512$ & $256$ & $2$ & LeakyReLU\\
			$3$ & tconv & $5 \times 5 \times 256$ & $128$ & $2$ & LeakyReLU\\
			$4$ & tconv & $5 \times 5 \times 128$ & $64$ & $2$ & LeakyReLU\\
			$5$ & tconv & $5 \times 5 \times 64$ & $1$ & $2$ & -\\
			\hline
		\end{tabular}
		\caption{}
		\label{table:appendix1.1.f}
	\end{subtable}
	\label{table:appendix1.1}
\end{table}

\newpage

\bibliographystyle{IEEEtran}
\bibliography{context_adaptive_neural_0,context_adaptive_neural_1}


\begin{IEEEbiography}[{\includegraphics[width=1in,height=1.25in,clip,keepaspectratio]{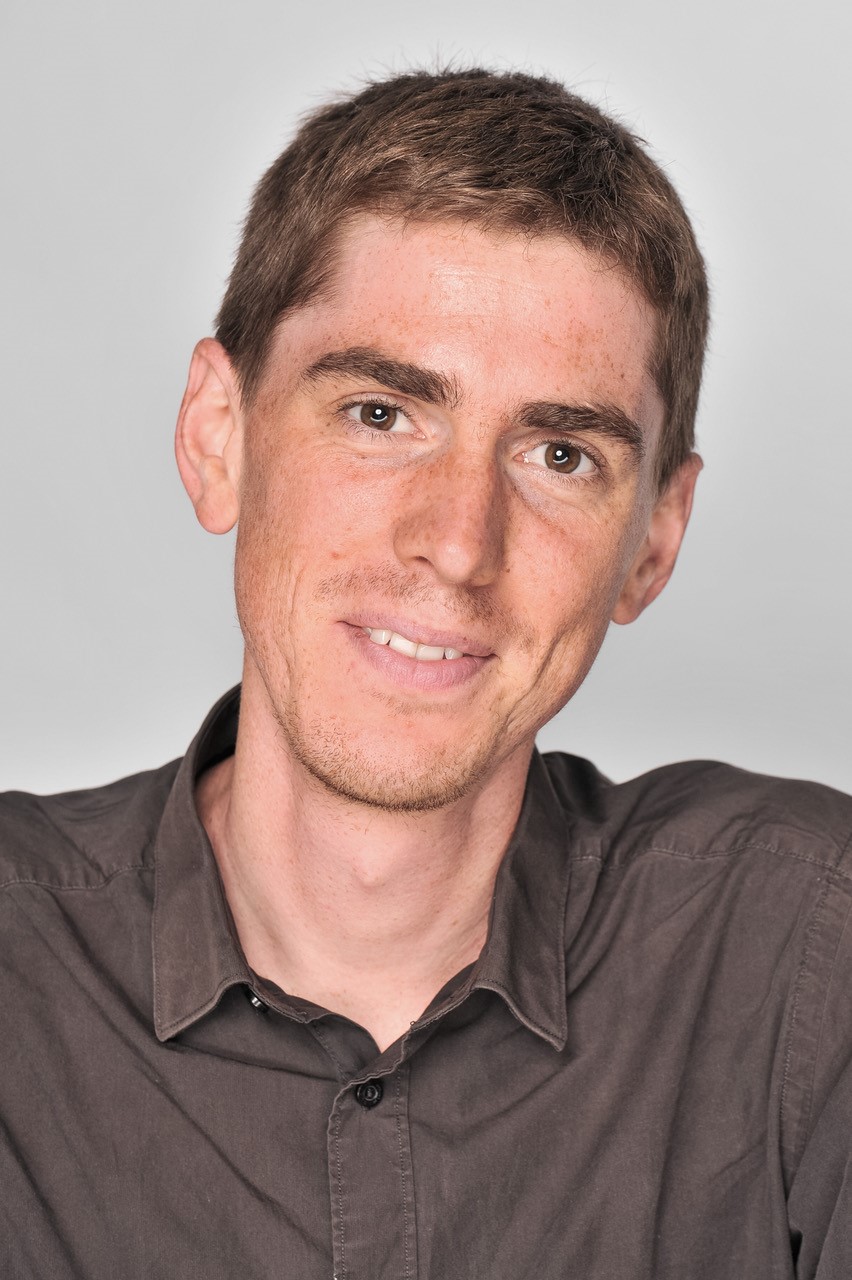}}]{Thierry Dumas}
received the Engineering degree and the Master degree from {\'E}cole Centrale Marseille, Marseille, France, in November 2014 and the Ph.D. degree from the University of Rennes I, France, in June 2019. He is now working as postdoctoral fellow at Interdigital, Rennes, France. His research interests include image and video compression.
\end{IEEEbiography}

\begin{minipage}{0.49\textwidth}
\begin{IEEEbiography}[{\includegraphics[width=1in,height=1.25in,clip,keepaspectratio]{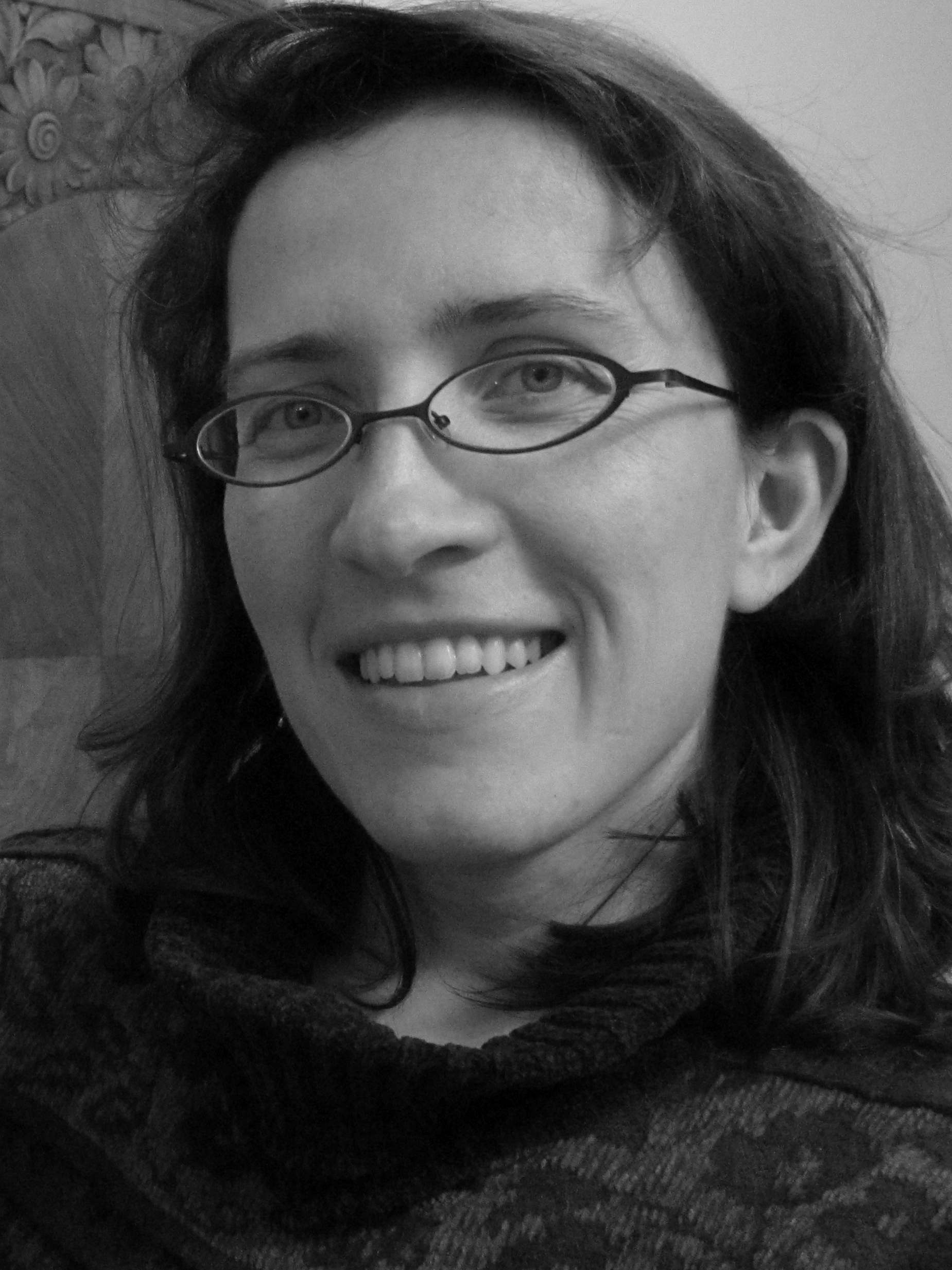}}]{Aline Roumy}
received the Engineering degree from {\'E}cole Nationale  Sup{\'e}rieure  de  l'{\'E}l{\'e}ctronique et de ses  Applications (ENSEA), Cergy, France, in 1996, the Master  degree in June 1997, and the Ph.D. degree from the University of Cergy-Pontoise, France, in September 2000. During 2000-2001, she was the recipient of a French Defense DGA/DRET postdoctoral fellowship and was a research associate at  Princeton University, Princeton, New Jersey. In November  2001, she joined INRIA Rennes, France. Her current research and study interests include the area of statistical signal processing, coding theory, and information theory.
\end{IEEEbiography}

\begin{IEEEbiography}[{\includegraphics[width=1in,height=1.25in,clip,keepaspectratio]{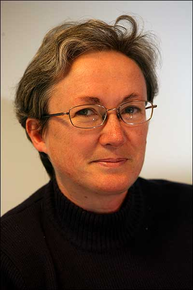}}]{Christine Guillemot} IEEE fellow, is Director of Research at INRIA, head of a research team dealing with image and video modeling, processing, coding and communication. She holds a Ph.D. degree from ENST (Ecole Nationale Sup{\'e}rieure des T{\'e}l{\'e}communications) Paris, and an Habilitation for Research Direction from the University of Rennes. From 1985 to October 1997, she has been with FRANCE TELECOM, where she has been involved in various projects in the area of image and video coding for TV, HDTV, and multimedia. From January 1990 to mid 1991, she has worked at Bellcore, NJ, USA, as a visiting scientist. Her research interests are signal and image processing, and in particular 2D and 3D image and video processing for various problems (compression, super-resolution, inpainting, classification).

She has served as Associate Editor for IEEE Trans. on Image Processing (from 2000 to 2003, and from 2014-2016), for IEEE Trans. on Circuits and Systems for Video Technology (from 2004 to 2006), and for IEEE Trans. on Signal Processing (2007-2009). She has served as senior member of the editorial board of the IEEE journal on selected topics in signal processing (2013-2015) and is currently senior area editor of IEEE Trans. on Image Processing.  
\end{IEEEbiography}
\end{minipage}

\end{document}